\documentclass{article}

\usepackage[main, final]{neurips_2026}

\usepackage[utf8]{inputenc}
\usepackage[T1]{fontenc}
\usepackage[hidelinks]{hyperref}
\usepackage{url}
\usepackage{booktabs}
\usepackage{amsfonts}
\usepackage{nicefrac}
\usepackage{microtype}
\usepackage{xcolor}

\usepackage{graphicx}
\usepackage{float}
\usepackage{array}
\usepackage{xspace}
\usepackage{multirow}
\usepackage{algorithm, algpseudocode}
\usepackage{stfloats}
\usepackage{natbib}
\usepackage{wrapfig}
\usepackage{subfig}
\usepackage{bbm}
\usepackage{caption}
\usepackage{bm}
\usepackage{enumitem}
\usepackage{amssymb, amsthm, amsmath, thm-restate, thmtools}
\usepackage{mathtools}
\usepackage{dsfont}
\numberwithin{equation}{section}

\newcommand{\alp}{\alpha}
\newcommand{\balp}{\bm{\alpha}}
\newcommand{\beps}{\bm{\eps}}
\newcommand{\bdel}{\bm{\del}}
\newcommand{\bc}{\mathbf{c}}
\newcommand{\bw}{\mathbf{w}}
\newcommand{\bv}{\mathbf{v}}

\newcommand{\bM}{\mathbf{M}}

\newcommand{\bT}{\mathbf{T}}
\newcommand{\lam}{\lambda}

\newcommand{\hatbalp}{\hat{\balp}}
\newcommand{\hatalp}{\hat{\alp}}
\newcommand{\hatpi}{\hat{\pi}}

\newcommand{\hatbw}{\hat{\bw}}
\newcommand{\hatJ}{\hat{J}}
\newcommand{\hatT}{\hat{T}}

\newcommand{\hatc}{\hat{c}}
\newcommand{\hatw}{\hat{w}}

\newcommand{\hatq}{\hat{q}}
\newcommand{\hatpii}{\hat{\pi}_i}

\newcommand{\mupi}{\mu_{\pi}}
\newcommand{\mupie}{\mu_{\pi_e}}
\newcommand{\mupii}{\mu_{\pi_i}}
\newcommand{\mupiavg}{\mu_{\pi_{\avg}}}
\newcommand{\mulog}{\mu_{\log}}
\newcommand{\pii}{\pi_i}
\newcommand{\pij}{\pi_j}
\newcommand{\pik}{\pi_k}
\newcommand{\pie}{\pi_e}
\newcommand{\avg}{\mathrm{avg}}
\newcommand{\piavg}{\pi_\avg}
\newcommand{\dm}{\mathrm{DM}}

\newcommand{\oips}{\mathrm{oIPS}}
\newcommand{\bips}{\mathrm{bIPS}}
\newcommand{\wips}{\mathrm{wIPS}}
\newcommand{\ips}{\mathrm{IPS}}

\newcommand{\drbips}{{\mathrm{DR}\text{-}\mathrm{bIPS}}}
\newcommand{\droips}{{\mathrm{DR}\text{-}\mathrm{oIPS}}}
\newcommand{\wipsclass}{\Gamma_\mathrm{wIPS}}

\newcommand{\drwipsclass}{\Gamma_\mathrm{DRwIPS}}
\newcommand{\supp}{\mathrm{supp}}
\newcommand{\expec}{\mathbb{E}}
\newcommand{\var}{\mathrm{Var}}
\newcommand{\cov}{\mathrm{Cov}}
\newcommand{\innp}[1]{\left\langle #1 \right\rangle}

\newcommand{\sbra}[1]{\left[ #1 \right]}
\newcommand{\rbra}[1]{\left( #1 \right)}

\newcommand{\ind}{\mathds{1}}

\newtheorem{lemma}{Lemma}[section]

\newtheorem{assumption}{Assumption}

\newtheorem{remark}{Remark}

\newcommand\numberthis{\addtocounter{equation}{1}\tag{\theequation}}

\def \A {\mathcal{A}}

\def \D {\mathcal{D}}
\def \F {\mathcal{F}}

\def \N {\mathcal{N}}

\def \R {\mathcal{R}}
\def \calS {\mathcal{S}}

\def \V {\mathcal{V}}
\def \eps {\epsilon}
\def \del {\delta}

\def \vw {\mathbf{w}}

\def \nat {\mathbb{N}}
\def \rr {\mathbb{R}}

\title{Variational Approach to Optimal IPS Estimator for Multi-logger Off-Policy Evaluation}

\author{Joon Suk Huh\thanks{These authors contributed equally to this work.} \\
Department of Computer Science\\
University of Wisconsin–Madison, USA \\
\texttt{joon@cs.wisc.edu} \\
\And
Junghoon Seo\footnotemark[1] \\
PIT IN Corp., Republic of Korea \\
\texttt{sjh@pitin-ev.com}
}

\begin{document}

\maketitle

\begin{abstract}
We study off-policy evaluation (OPE) in contextual bandits with data collected from multiple logging policies. Inverse propensity scoring (IPS) is a standard approach to OPE and extends naturally to the multi-logger setting. However, as highlighted by~\cite{agarwal2017effective}, neither balanced nor weighted IPS consistently outperforms the other in this setting. We resolve this dilemma by deriving an optimal IPS estimator with sample-dependent weights that minimize variance subject to unbiasedness. Using a variational calculus approach, we obtain closed-form optimal weights, yielding an estimator that is unbiased and achieves asymptotically optimal variance within a generalized weighted-IPS estimator class. Instead of a reward model, the estimator needs only a $K\times K$ matrix and a $K$-vector of moments, where $K$ is the number of logging policies, and we quantify how much efficiency this costs relative to the semiparametric oracle. Experiments on benchmark datasets show that our estimator generally improves on existing multi-logger IPS methods. We also extend our estimator to a doubly robust form by incorporating a conditional reward estimator. We compare the resulting DR extension with the semiparametrically efficient DR estimator of~\citet{kallus2021optimal}, both theoretically and empirically. We show that the two estimators achieve the same asymptotic variance when the conditional reward estimator is consistent, while our estimator can retain semiparametric efficiency under certain forms of reward misspecification. Empirically, the proposed DR estimator often achieves lower estimation error than the existing DR baseline.
\end{abstract}

\section{Introduction}\label{sec:intro}

Off-policy evaluation (OPE) is a core tool for data-driven decision making, with applications in recommender systems, healthcare, and advertising. The goal is to estimate the value of an evaluation policy using data collected by one or more different logging policies. While the single-logger setting is well understood, many practical datasets are collected from multiple logging policies, for example through heterogeneous systems, parallel deployments, or sequential changes to a production policy~\citep{agarwal2017effective,saito2021open,jung2024efficient}. This multi-logger structure creates a basic statistical challenge: different strata can have very different overlaps with the evaluation policy, so combining them poorly can lead to large variance.

Inverse propensity scoring (IPS) is a standard approach to OPE~\citep{rosenbaum1983central,precup2000eligibility,bottou2013counter}. In the single-logger case, IPS reweights each observed reward by the propensity ratio between the evaluation and logging policies. In the multi-logger case, however, there is an additional design choice: one must decide how much each logger should contribute to the final estimate. Existing approaches answer this question in different ways. Naive IPS treats all logged samples with their own propensity ratios, balanced IPS uses the sample-size-weighted average logging policy, and weighted IPS chooses logger-level weights based on estimated variances~\citep{agarwal2017effective}. As observed by~\citet{agarwal2017effective}, the relative performance of balanced and weighted IPS is highly instance dependent: neither consistently dominates the other across problem instances, and naive IPS is dominated in their analysis. \citet{kallus2021optimal} later characterized the semiparametrically efficient estimator over a broader class, which serves as our benchmark.

We revisit this problem from a variational perspective. Instead of choosing among a fixed set of IPS estimators, we consider a generalized weighted-IPS class $\wipsclass$ in which each sample may receive a weight depending on the logged tuple and its logger identity. That is, the weights are functions of the state, action, reward, and logger. We impose constraints that ensure unbiasedness and then directly minimize the variance functional over all feasible weight functions. This yields a convex variance minimization problem whose solution admits a closed-form characterization. The resulting estimator, denoted $\hatJ_\oips$, can be viewed as a balanced-IPS estimator augmented with an optimal control-variate term built from the logging propensities. This construction does not require the conditional mean reward. Instead, it estimates a $K\times K$ matrix $\bT$ and a vector $\bc\in\rr^K$ of moments (\S\ref{sec:optimalIPS}).

This perspective clarifies the apparent instance-dependence of existing IPS estimators. The optimal weights depend on the interaction among the evaluation policy, the logging policies, the reward distribution, and the stratum sizes. Thus, fixed IPS rules such as naive IPS, balanced IPS, and weighted IPS can each perform well on some instances and poorly on others. By optimizing over a class containing these IPS baselines, $\hatJ_\oips$ adapts to the instance and achieves the smallest asymptotic variance within $\wipsclass$. We prove that the corresponding feasible cross-fitted implementation is unbiased when logging policies are known and is asymptotically normal with the optimal variance.%

We also relate our estimator to the semiparametric efficiency theory of~\citet{kallus2021optimal}. Their framework shows that the most efficient estimator uses the conditional mean reward as a control variate. Our IPS estimator can be viewed as a restricted, reward model-free version of this idea: instead of estimating the conditional mean reward directly, it constructs the best control variate available from the logging policies. When this logger-induced control variate is rich enough to represent the relevant reward structure, our estimator asymptotically recovers the efficient oracle. Otherwise, its excess variance is determined by the part of the reward structure that cannot be represented by the logging policies.

Finally, we extend our variational approach to a doubly robust estimator, denoted $\hatJ_\droips$, by incorporating a reward model. This extension keeps the same idea as our IPS estimator: it uses the logging policies to choose an additional variance-reducing correction. This makes it directly comparable to the doubly robust balanced IPS estimator $\hatJ_\drbips$ of~\citet{kallus2021optimal}. When the reward model is correctly specified, both estimators attain the same asymptotic variance. When the reward model is misspecified, however, $\hatJ_\droips$ can remain efficient if the misspecification is representable by the logging policies, whereas $\hatJ_\drbips$ need not. This gives our DR extension an additional robustness advantage.

\paragraph{Summary of Contributions.} We make the following contributions:
\begin{enumerate}[leftmargin=*]
    \item \textbf{A variationally optimal IPS estimator for multi-logger OPE.} We introduce a generalized weighted-IPS class $\wipsclass$ and derive the variance-minimizing unbiased estimator within this class. The resulting estimator $\hatJ_\oips$ has closed-form sample-dependent weights and estimates $\bT$ and $\bc$ in place of a reward model.

    \item \textbf{Closed-form characterization, feasible estimation, and projection interpretation.} We show that the optimal weights are characterized by a linear system $\bT\balp=\bc$, whose components can be estimated from data, leading to a practical cross-fitted implementation. We further show that this system has a geometric interpretation: it projects the reward-weighted evaluation policy onto the span of the logging policies, and the resulting projection residual exactly determines the variance gap to the semiparametrically efficient oracle.

    \item \textbf{Doubly robust extension and comparison with efficient DR estimators.} We extend $\hatJ_\oips$ to a doubly robust estimator $\hatJ_\droips$ by applying the same variational construction to residuals based on $\hatq$. We show that $\hatJ_\droips$ matches the asymptotic variance of $\hatJ_\drbips$ when $\hatq$ is consistent, and can retain the oracle variance under certain structured misspecifications of $\hatq$.

    \item \textbf{Empirical validation.} Across benchmark datasets and a broad range of logger mixtures, our IPS estimator generally improves over existing multi-logger IPS baselines. The doubly robust extension also often achieves lower estimation error than the semiparametrically efficient baseline.
\end{enumerate}

\paragraph{Related Work.}
Appendix~\ref{app:related-work} discusses related work on OPE with multiple loggers~\citep{he2019off,chen2020infinite}, connections to multiple importance sampling~\citep{veach1995optimally,elvira2019generalized,kondapaneni2019optimal}, and broader variance-reduction methods for IPS.

\section{Preliminaries}\label{sec:prelim}
This section introduces the problem setup and reviews baseline estimators used in our experiments.

\paragraph{Notation.}
For $n \in \nat$, let $[n] \coloneq \{1, \dots, n\}$. We use boldface for finite-dimensional vectors and matrices, e.g., $\bv \in \rr^n$ and $\bM \in \rr^{n \times n}$. Their entries are written in plain font (e.g., $v_i$ and $M_{ij}$), and we likewise use plain symbols for scalar components such as $\alp_i$, $c_i$, and $w_i$. The symbol $\ind[\,\cdot\,]$ denotes the indicator function.

Let $(\Omega,\mathcal F,\mu)$ be a $\sigma$-finite measure space and let $f,g \in L^2(\mu)$.
We write $\langle\cdot,\cdot\rangle_\mu$ for the $L^2(\mu)$ inner product:
\begin{align*}
    \innp{f,g}_\mu
    \coloneq \int_\Omega f(\omega)\,g(\omega)\,d\mu(\omega).
\end{align*}
The associated $L^2(\mu)$-norm is $\|f\|_{L^2(\mu)} \coloneq \innp{f,f}_\mu^{1/2}$. We use the same notation $\innp{f,g}_\mu$ whenever $fg$ is $\mu$-integrable, even if $f$ or $g$ is not in $L^2(\mu)$.

Throughout the paper, we fix a $\sigma$-finite measure space $(\Omega,\F,\mu)$, where $\Omega\coloneq \calS\times\A\times\R$, with $\calS$, $\A$, and $\R$ denoting the sets of states, actions, and rewards, respectively, as defined in \S\ref{subsec:problem-setting}. For $\omega=(s,a,r)\in\Omega$, let $\rho_r(\omega)\coloneq r$ denote the projection onto the reward coordinate; $R$ denotes the corresponding random variable. Here, $\F$ is a suitable $\sigma$-algebra and $\mu$ is a fixed reference measure. For brevity, we write $L^2\coloneq L^2(\mu,\Omega)$. For any measure $\nu \ll \mu$, we write $L^2(\nu)$ for the space of square-integrable functions with respect to $\nu$.
When $\Omega$ is a topological space, we denote by $\supp(\mu)$ the support of a measure $\mu$, and by $\supp(f)$ the support of a measurable real-valued function $f$. For a policy $\pi$, we write $\supp(\pi(\cdot|s))\coloneq\mathrm{cl}\rbra{\{a\in\A:\pi(a\mid s)>0\}}$.
For measurable functions $f$ and $g$ with
$g\neq0$ wherever $f\neq0$, we define the ratio $f/g$ pointwise, with the convention that $0/0 = 0$.

\subsection{Problem Setting}\label{subsec:problem-setting}
We study OPE in contextual bandits, following the multi-logger setup in~\citep{agarwal2017effective, he2019off, kallus2021optimal}.
A policy $\pi$ maps a state (i.e., context) $s\in\calS$ to a distribution over actions in $\A$, written as $\pi(a\mid s)$.

An environment is defined by the following \emph{unknown} distributions: a context distribution $p_s(s)$ and reward distributions $p_r(r\mid s,a)$ over $\R\subseteq\rr$ for each context-action pair $(s,a)\in\calS\times\A$. Interaction proceeds as follows: the environment samples $S\sim p_s$, the actor samples $A\sim\pi(\,\cdot\mid S)$, and the environment then samples the reward $R\sim p_r(\,\cdot\mid S,\,A)$. This induces a joint distribution $\mupi$ on $\calS\times\A\times\R$, whose density is $p_s(s)\,\pi(a|s)\, p_r(r|s,a)$ with respect to the dominating measure $\mu(ds,da,dr)=\mu_s(ds)\,\mu_a(da)\,\mu_r(dr\mid s,a)$, where $\mu_r(\cdot\mid s,a)$ is a reference kernel on $\R$. Equivalently, $\mupi(ds,da,dr)=P_S(ds)\,\pi(da\mid s)\,P_R(dr\mid s,a)$, where $P_S(ds)=p_s(s)\,\mu_s(ds)$, $\pi(da\mid s)=\pi(a\mid s)\,\mu_a(da)$, and $P_R(dr\mid s,a)=p_r(r\mid s,a)\,\mu_r(dr\mid s,a)$.
The expected reward of the policy $\pi$, denoted $J(\pi)$, is
\begin{align*} 
    J(\pi) \coloneq \expec_{\mupi}[R], 
\end{align*}
where $\expec_{\mupi}$ denotes expectation with respect to $\mupi$.

The goal of OPE is to estimate the policy value $J(\pie)$ of an evaluation policy $\pie$ from logged data. In the single-logger case, we observe i.i.d. samples $\D\coloneq\big\{(S_j,A_j,R_j)\big\}_{j=1}^n$ drawn from $\mupi$, where $\pi$ is the logging policy and typically $\pi\neq\,\pie$. In this work, we consider stratified OPE with multiple logging policies $(\pii)_{i\in[K]}$. The full dataset is $\D\coloneq\{\D_i\}_{i\in[K]}$, where the strata are independent and each stratum $\D_i$ consists of $n_i$ i.i.d. samples from $\mupii$, namely $\D_i\coloneq\big\{(S_{ij},A_{ij},R_{ij})\big\}_{j=1}^{n_i}$, where $S_{ij}\sim p_s$, $A_{ij}\sim\pii(\cdot\mid S_{ij})$, and $R_{ij}\sim p_r(\cdot\mid S_{ij},A_{ij})$.

\subsection{Existing Inverse Propensity Scoring (IPS)-based Approaches}
\paragraph{Naive IPS.}
Naive IPS reweights each observed reward by the propensity ratio between the evaluation and logging policies:
\begin{align*}
    \hatJ_{\ips}(\pie;\D) = \frac{1}{n} \sum_{i=1}^K \sum_{(S,A,R)\in\D_i} R\,\frac{\pie(A\mid S)}{\pii(A\mid S)},
    \numberthis\label{eq:IPS}
\end{align*}
where $n\coloneq\sum_{i=1}^K n_i$ is the total number of samples.

\paragraph{Balanced IPS.}
\citet{agarwal2017effective} proposed a multiple importance sampling (MIS)-inspired estimator that uses the averaged logging policy as inverse propensity score:
\begin{align*}
    \hatJ_{\bips}(\pie;\D) = \frac{1}{n} \sum_{i=1}^K \sum_{(S,A,R)\in\D_i}R\,\frac{\pie(A\mid S)}{\piavg(A\mid S)},
    \numberthis\label{eq:bIPS}
\end{align*}
where $\piavg\coloneq\frac{1}{n}\sum_{i=1}^K n_i\,\pii$ and $n_i\coloneq|\D_i|$.

\paragraph{Weighted IPS.}
\citet{agarwal2017effective} also introduced weighted IPS, which combines logger-specific IPS estimates using variance-based weights:
\begin{align*}
    \hatJ_{\wips}(\pie;\D) = \sum^K_{i=1} \hat{\gamma}_i\, \sum_{(S,A,R)\in\D_i} R\,\frac{\pie(A\mid S)}{\pii(A\mid S)},
    \numberthis\label{eq:wIPS}
\end{align*}
where $\hat{\gamma}_i \coloneq \rbra{\widehat{\sigma}_r^2(\pie||\pii)  \sum^K_{j=1} \frac{n_j}{\widehat{\sigma}_r^2(\pie||\pij)} }^{-1}$ 
and $\hat{\sigma}^2_r(\pie||\pii)$ is a positive consistent estimator of $\var_{\mupii}\!\sbra{R\, \frac{\pie(A\mid S)}{\pii(A\mid S)}}$, which quantifies the divergence between the target and logging policies.

\paragraph{Doubly Robust Balanced IPS.}
\citet{kallus2021optimal} proposed a doubly robust version of $\hatJ_\bips$ using a conditional mean reward estimator $\hatq$:
\begin{align*}
    \hatJ_{\drbips}(\pie;\D) = \frac{1}{n}\sum_{i=1}^K \sum_{(S,A,R)\in\D_i}\sbra{\frac{\pie(A\mid S)}{\piavg(A\mid S)}\big(R - \hatq(S,A)\big) + \hatq(S,\pie)},
    \numberthis\label{eq:dr-bIPS}
\end{align*}
where $\hatq(S,\pie)\coloneq \int_{\A}\hatq(S,a)\pie(a\mid S)\,d\mu_a(a)$. When the logging policies are known and $\hatq$ consistently estimates the conditional mean reward $q(s,a)\coloneq\expec[R\mid S=s,A=a]$, $\hatJ_{\drbips}$ attains the semiparametric efficiency bound within the estimator class studied by~\citet{kallus2021optimal}, under their regularity conditions. In practice, $\hatq$ is estimated by cross-fitting: the reward model is trained on one part of the data and evaluated on held-out samples. The same cross-fitting strategy for nuisance quantities is also used in $\hatJ_{\wips}$.

\section{Variational Optimization Approach}\label{sec:optimalIPS}
In this section, we introduce our main estimator, which is derived by optimizing a convex variance functional and is asymptotically optimal within the generalized weighted-IPS class denoted by $\wipsclass$.

\subsection{Generalized Weighted-IPS Estimator Class}
Let $\bw\coloneq (w_i)_{i\in[K]}$, where each $w_i:\calS\times\A\times\R\to\rr$ is a sample-dependent weight function. The generalized weighted-IPS estimator class $\wipsclass$ is
\begin{align*}
\boxed{
    \hatJ(\pie,\bw;\D)\coloneq\sum^K_{i=1}\frac{1}{n_i} \sum_{(S,A,R)\in\D_i} w_i(S,A,R)\,\frac{\pie(A\mid S)}{\pii(A\mid S)}
    \numberthis\label{eq:general-weighted-ips}
}
\end{align*}
Estimators in $\wipsclass$ are unbiased when the weights satisfy the following conditions for every $\omega=(s,a,r)$:
\begin{align*}
    \textbf{Normalization:}\ \sum_{i=1}^K w_i(\omega) = \rho_r(\omega),\quad \textbf{Dominance:}\ \pii(a|s)=0\implies w_i(s,a,r)=0.
\end{align*}

\begin{restatable}{theorem}{Junbiased}\emph{[Unbiasedness]}
    For any weights $\bw\in(L^2)^K$ that satisfy the normalization and dominance conditions and for which each weighted score $w_i(S,A,R)\,\pie(A\mid S)/\pii(A\mid S)$ is $\mupii$-integrable, $\hatJ(\pie,\bw;\D)$ is unbiased, that is, $J(\pie)= \expec[\hatJ(\pie,\bw;\D)]$, where the expectation is over the sampling randomness of logged data $\D\coloneq(\D_i)_{i\in [K]}$.
    \label{thm:unbiasedness}
\end{restatable}
In \S\ref{subsec:comparison}, we compare $\wipsclass$ with the broader class of~\citet{kallus2021optimal}, whose efficient estimator requires an estimate of the conditional mean reward.

\paragraph{Variationally Optimized Weights.}
Since $\hatJ(\pie,\bw;\D)$ is a linear functional of $\bw$, its variance is convex. We optimize this variance functional directly over feasible $\bw$ satisfying the normalization and dominance constraints. The resulting optimal weight vector $\bw^\star\coloneq(w_i^\star)_{i\in [K]}$, whose optimality is established in Theorem~\ref{thm:optimalIPS}, takes the form
\begin{align*}
    w^\star_i(s,a,r)=
    \frac{\alp^\star_i\pii(a|s)}{\pie(a|s)}
    +\frac{n_i\,\pii(a|s)}{n\,\piavg(a|s)}
    \rbra{
        r-\frac{\sum_{j=1}^K\!\alp^\star_j\,\pij(a|s)}{\pie(a|s)}
    }
    \numberthis\label{eq:w-star}
\end{align*}
where $\balp^\star\coloneq(\alp^\star_i)_{i\in[K]}$ is a solution of $\bT\balp=\bc$, with $\bT\in\rr^{K\times K}$ and $\bc\in\rr^K$ whose components are given by
\begin{align*}
    T_{ij}\coloneq \expec_{\mupii}\!\sbra{\frac{\pij(A\mid S)}{\piavg(A\mid S)}},\ \  
    c_i\coloneq \expec_{\mupii}\!\sbra{R\,\frac{\pie(A\mid S)}{\piavg(A\mid S)}}
    \numberthis\label{eq:Tc}
\end{align*}
We call $\bT$ the \emph{technique matrix} and $\bc$ the \emph{contribution vector}. Estimating $\bT$ and $\bc$ thus replaces estimating the conditional mean reward function. We later show in this section that the equation $\bT\balp=\bc$ can be interpreted as the normal equation for an orthogonal projection.

\paragraph{Control-Variate Form.} 
Plugging the expression for $\bw^\star$ in~\eqref{eq:w-star} into $\hatJ(\pie,\bw^\star;\D)$ in~\eqref{eq:general-weighted-ips} yields the following form of the variationally optimal IPS estimator \footnote{With a slight abuse of notation, we write $\hatJ(\pie,\balp^\star;\D)=\hatJ(\pie,\bw^\star;\D)$.}:
\begin{align*}
\boxed{
    \hatJ(\pie,\balp^\star;\D)
    = \frac{1}{n}\sum_{i=1}^K\sum_{(S,A,R)\in\D_i}
    \Bigg[
       \,\frac{R\,\pie(A\mid S)}{\pi_\avg(A\mid S)}+\sum_{k=1}^K\alp^\star_k-\frac{g_{\balp^\star}(S,A)}{\piavg(A\mid S)}
    \Bigg]
    \numberthis\label{eq:hatJ-control-variate-form}
}
\end{align*}
where $g_{\balp^\star}(s,a)\coloneq \sum_{k\in [K]}\alp^\star_k\,\pik(a|s)$. We refer to this as the control-variate form: the term involving $\sum_k \alp_k^\star-g_{\balp^\star}/\piavg$ is mean-zero and therefore acts as a control variate.

\paragraph{Geometric View.} 
Let $q(s,a)\coloneq\expec[R\,|\,S=s,A=a]$ denote the conditional mean reward. Then, we can view $g_{\balp^\star}$ as the orthogonal projection of $q\,\pie$ onto the finite-dimensional space spanned by $\{\pii\}_{i\in [K]}$.
Specifically, consider $L^2(\mulog,\calS\times\A)$, where $\mulog$ is given by
\begin{align*}
    d\mulog(s,a) \coloneq \frac{p_s(s)}{\piavg(a|s)}\,d\mu_{s,a}(s,a),
\end{align*}
with $\mu_{s,a}=\mu_s\otimes\mu_a$ denoting the base measure over $\calS\times\A$. Equip $L^2(\calS\times\A,\mulog)$ with the inner product $\innp{\cdot,\cdot}_{\mulog}$. Then, one can rewrite~\eqref{eq:Tc} in terms of $\innp{\cdot,\cdot}_{\mulog}$ as
\begin{align*}
    T_{ij}=\big\langle\pii,\pij\big\rangle_{\mulog},\quad c_i=\big\langle q\,\pie,\pii \big\rangle_{\mulog}.
\end{align*}
Thus, $\bT\balp^\star=\bc$ if and only if $\big\langle q\,\pie-\sum_j \alp^\star_j \pij,\pii\big\rangle_{\mulog}=0$ for all $i\in[K]$. This shows that $g_{\balp^\star}$ is the projection of $q\,\pie$ onto $\mathrm{span}\big(\{\pii\}_{i\in[K]}\big)$, with respect to the norm induced by $\innp{\cdot,\cdot}_{\mulog}$.

We formally state this observation in Lemma~\ref{lem:self-consistency-as-projection}. Later, in \S\ref{subsec:comparison}, we relate the squared $\|\cdot\|_{\mulog}$ norm of the residual component of $q\,\pie$ orthogonal to $\mathrm{span}\big(\{\pii\}_{i\in[K]}\big)$ to the excess variance of our estimator relative to the semiparametrically efficient oracle of~\citet{kallus2021optimal}.

\subsection{Feasible Variationally Optimal Estimator $\hatJ_\oips$}\label{subsec:feasible-ips}
In this section, we derive a \emph{feasible} variationally optimal estimator $\hatJ_\oips$. Specifically, we employ the logger-wise cross-fitting strategy~\citep{zheng2011cross,chernozhukov2018double,kallus2021optimal}: one part of the data is used to estimate the nuisance parameter $\balp$ by estimating $\bT$ and $\bc$, while the other part is used to compute the estimator~\eqref{eq:hatJ-control-variate-form}.

\paragraph{Computing $\hatJ_\oips$.}
For each $i \in [K]$, let $\big\{\D^{(z)}_i\big\}_{z \in [Z]}$ be a $Z$-fold partition ($Z\geq2$) of $\D_i$ into nonempty folds chosen independently of the observed values, with $\D^{(-z)}_i\coloneq \D_i\setminus \D_i^{(z)}$ and $n_i^{(-z)}\coloneq \big|\D_i^{(-z)}\big|$.
Then, for each fold $z\in [Z]$, estimate $\bT,\bc$ by 
\begin{align*}
    \hatT_{ij}^{(z)} \coloneq \frac{1}{n_i^{(-z)}}\sum_{(S,A,R)\in \D_i^{(-z)}} \frac{\pij(A\mid S)}{\piavg^{(-z)}(A\mid S)},\quad
    \hatc_i^{(z)}\coloneq \frac{1}{n_i^{(-z)} }\sum_{(S,A,R)\in \D_i^{(-z)}}\frac{R\,\pie(A\mid S)}{\piavg^{(-z)}(A\mid S)}
    \numberthis\label{eq:Tchat}
\end{align*}
where $\piavg^{(-z)}\coloneq \frac{1}{n^{(-z)}}\sum_{i\in [K]} n_i^{(-z)} \pii$ with $n^{(-z)} \coloneq \sum_{i\in [K]}n_i^{(-z)}$.
Then, we estimate $\balp^\star$ with 
$\hat{\balp}^{(z)} \coloneq \big(\hat{\bT}^{(z)}\big)^\dagger\hat{\bc}^{(z)}$, where $\dagger$ denotes the Moore--Penrose pseudoinverse.

Let $n_i^{(z)}\coloneq \big|\D_i^{(z)}\big|$ and $n^{(z)}\coloneq \sum_{i\in [K]} n^{(z)}_i$. Then, the feasible variationally optimal estimator is
\begin{align*}
\boxed{
    \hatJ_\oips(\pie;\D)\coloneq\frac{1}{n}\sum_{z=1}^Z n^{(z)}\hatJ\big(\pie,\hat{\balp}^{(z)};\D^{(z)}\big)
    \numberthis\label{eq:optimalIPS}
}
\end{align*}
where $\hatJ\rbra{\pie,\hat{\balp}^{(z)};\D^{(z)}}$ is~\eqref{eq:hatJ-control-variate-form} evaluated with $\hat{\balp}^{(z)}$ and $\D^{(z)}\coloneq\big\{\D_i^{(z)}\big\}_{i\in [K]}$ in place of $\balp^\star$ and $\D$. Note that the quantity $\piavg$ in~\eqref{eq:hatJ-control-variate-form} depends on $\D$ through $n_i/n$, so it is also replaced with $\piavg^{(z)}\coloneq \frac{1}{n^{(z)}}\sum_{i\in [K]} n_i^{(z)} \pii$.

\paragraph{When Logging Policies are Unknown.}
When the logging policies are unknown, we plug consistent estimators of $\pi_i(a|s)$ and $\piavg(a|s)$ into the above procedure. The estimation procedures are given in Appendix~\ref{app:experiment-results}.

In Theorem~\ref{thm:optimalIPSasymptotics}, we show that $\hatJ_\oips$ is unbiased and asymptotically normal with the optimal asymptotic variance $n\,\var[\hatJ(\pie,\bw^\star;\D)]$ when the logging policies are known; Remark~\ref{rem:estimated} discusses estimated logging policies.

\section{Theoretical Analysis}\label{sec:analysis}
In this section, we analyze $\hatJ_\oips$ introduced in \S\ref{sec:optimalIPS} and prove its asymptotic variance optimality within the generalized weighted-IPS class~\eqref{eq:general-weighted-ips}. We then compare $\hatJ_\oips$ with other estimators, including the estimator $\hatJ_\drbips$~\eqref{eq:dr-bIPS}, which is asymptotically optimal within a class containing the three IPS baselines~\citep{kallus2021optimal} but requires an estimator of the conditional mean reward. We characterize the variance gap between $\hatJ_\oips$ and $\hatJ_\drbips$ in terms of the residual from projecting the evaluation policy $\pie$, weighted by the conditional mean reward $q(s,a)$, namely $q\,\pie$, onto the logger basis space.

For simplicity of the analysis, we make the following assumptions.
\begin{assumption}[Regularity]\label{as:regularity}
\leavevmode
\begin{enumerate}[leftmargin=*,label=(\roman*)]
    \item All logging and evaluation policies have strictly positive densities over the action space at every state in the support of $p_s$.\footnote{Consequently, we drop the dominance condition in the variance minimization problem, but our solution still satisfies it.}
    \item The weighted scores $w_i(S,A,R)\,\pie(A\mid S)/\pii(A\mid S)$ have finite second moments under $\mupii$, and the optimal weights in~\eqref{eq:w-star} are square-integrable.
    \item For the feasible estimator, $K$ and $Z$ are fixed and the folds of each stratum have equal sizes up to rounding. For every $i\in[K]$ and $z\in[Z]$, the summand of~\eqref{eq:hatJ-control-variate-form} computed with $\hat\balp^{(z)}$ converges in probability to the summand computed with $\balp^\star$ in $L^2(\mupii)$, and both summands have finite second moments, including over the randomness of the training folds.
\end{enumerate}
\end{assumption}
\begin{remark}\label{rem:deterministic}
If $\pie$ has zeros, as for the deterministic evaluation policy in \S\ref{sec:experiment}, Assumption~\ref{as:regularity}(i) fails: the weights~\eqref{eq:w-star} are undefined where $\pie(a\mid s)=0$, and the control-variate form~\eqref{eq:hatJ-control-variate-form} is no longer a member of $\wipsclass$. The control-variate form remains unbiased for every $\balp$, and any solution of $\bT\balp=\bc$ still minimizes its variance over $\balp\in\rr^K$. The comparisons in \S\ref{subsec:comparison} extend to such $\pie$ by continuity, by applying them to $\pie^\epsilon\coloneq(1-\epsilon)\pie+\epsilon\pi_u$, where $\pi_u$ is the uniform policy, and letting $\epsilon\to0$.
\end{remark}

\subsection{Derivation of Optimal Weights}
Let $\hatJ(\pie,\bw;\D)$ be a generalized weighted-IPS estimator~\eqref{eq:general-weighted-ips}. Then, $\var\big[\hatJ(\pie,\bw;\D)\big]$ is a convex functional of $\bw$, so the variance can be optimized directly over the feasible weights. The main convex program of interest is
\begin{align*}
    & \underset{\bw\in(L^2)^K}{\text{minimize}}
    & &\hspace{-0.5in}\var\big[\hatJ(\pie,\bw;\D)\big]\numberthis\label{eq:var-min-problem}\\
    & \text{subject to}
    & &\hspace{-0.5in}\sum^K_{i=1} w_i(\omega)=\rho_r(\omega), \quad\forall \omega\in\supp\,(\mupie),%
\end{align*}

\begin{restatable}{theorem}{optimalIPS} 
    The weight vector $\bw^\star$ given in~\eqref{eq:w-star} solves the variance minimization problem~\eqref{eq:var-min-problem}.
    \label{thm:optimalIPS}
\end{restatable}
The proof of this theorem is given in Appendix~\ref{app:subsec:proof:thm:optimalIPS}.

By Theorem~\ref{thm:optimalIPS}, computing $\hatJ_\oips$ only requires estimating $\bT$ and $\bc$, whose $i$-th rows are averages over stratum $i$, and then solving the $K\times K$ system $\bT\balp=\bc$ for each fold. Assuming each propensity evaluation takes constant time, accumulating the moments takes $O(nK)$ time and each solve takes $O(K^3)$, so $Z$-fold cross-fitting costs $O(Z(nK+K^3))$, which is $O(nK+K^3)$ for fixed $Z$. If the per-fold moments are accumulated in one pass over the data, the extra memory is $O(ZK^2)$, and a second pass evaluates the estimator; storing every cross-logger propensity instead takes $O(nK)$. These counts exclude fitting nuisance models such as propensity models. Appendix~\ref{app:complexity} reports the corresponding runtimes.

\subsection{Unbiasedness and Asymptotic Variance Optimality of $\hatJ_{\oips}$ within $\wipsclass$}
We next analyze the feasible cross-fitted estimator $\hatJ_{\oips}$. Specifically, we prove its unbiasedness and asymptotic variance optimality within the generalized weighted-IPS estimator class $\wipsclass$. We follow the multi-logger asymptotic regime of~\citet{kallus2021optimal}: for each logger, the number of observations is $n'_i=m n_i$, and we consider the limit $m\to\infty$.

\begin{restatable}{theorem}{optimalIPSasymptotics}
    Let $n'\coloneq \sum_{i=1}^K n'_i$ with $n'_i\coloneq m n_i$. Under Assumption~\ref{as:regularity}, when the logging policies are known, $\hatJ_\oips$ is unbiased, and
    \begin{align*}
        \sqrt{n'}\rbra{\hatJ_\oips(\pie;\D) - J(\pie)} \overset{d}{\to} \N\rbra{0, n\,\var[\hatJ(\pie,\vw^\star;\D)]}
    \end{align*}
    as $m\to\infty$, where $n\coloneq \sum_{i=1}^K n_i$ and $\var[\hatJ(\pie,\vw^\star;\D)]$ is computed at the base stratum sizes $(n_i)_{i=1}^K$.
\label{thm:optimalIPSasymptotics}
\end{restatable}
That is, $\hatJ_\oips$ is consistent and asymptotically normal, with the optimal asymptotic variance $n\,\var[\hatJ(\pie,\vw^\star;\D)]$ achievable within the generalized weighted-IPS estimator class $\wipsclass$. The proof of Theorem~\ref{thm:optimalIPSasymptotics} is given in Appendix~\ref{proof:thm:optimalIPSasymptotics}.
\begin{remark}[Estimated logging policies]\label{rem:estimated}
When the logging policies are unknown, we plug in estimates fitted on the training folds. Under mild regularity conditions, $\hatJ_\oips$ remains consistent if these estimates are consistent, but its limiting distribution generally differs from that in Theorem~\ref{thm:optimalIPSasymptotics}: the score of $\hatJ_\oips$ is not orthogonal to the propensities, so propensity estimation error enters at first order, as for IPS with estimated propensities~\citep{hirano2003efficient}. The limit in Theorem~\ref{thm:optimalIPSasymptotics} still holds if the resulting fold-wise bias is $o_p(n^{-1/2})$, for example when $q\,\pie\in\mathrm{span}(\{\pii\}_{i\in[K]})$.
\end{remark}

\subsection{Theoretical Comparison with Other Estimators}\label{subsec:comparison}
We provide a theoretical comparison of the estimators in terms of their asymptotic variance.

\paragraph{Comparison with $\hatJ_\ips$, $\hatJ_\bips$ and $\hatJ_\wips$.} $\wipsclass$ contains the naive IPS estimator ($\hatJ_\ips$), the balanced IPS estimator ($\hatJ_\bips$), and the weighted IPS estimator ($\hatJ_\wips$). Setting the weights in~\eqref{eq:general-weighted-ips} to
\begin{align*}
    w_i^{\ips}=\frac{n_i}{n}r,\qquad
    w_i^{\bips}=\frac{n_i}{n}\frac{\pi_i}{\piavg}r,\qquad
    w_i^{\wips}=n_i\hat\gamma_i r
\end{align*}
recovers~\eqref{eq:IPS}, \eqref{eq:bIPS}, and~\eqref{eq:wIPS}, respectively. Each choice satisfies the normalization constraint because $\sum_i n_i/n=1$, $\sum_i(n_i/n)\pii/\piavg=1$, and $\sum_i n_i\hat\gamma_i=1$. The same identities hold for the cross-fitted version of $\hatJ_\wips$, conditionally on the training folds, with the held-out stratum sizes in place of $n_i$. Hence, with known propensities, our asymptotic efficiency result in Theorem~\ref{thm:optimalIPSasymptotics} implies that, for every problem instance,
\begin{align*}
    \mathrm{AVar}\big[\hatJ_\oips(\pie;\D)\big] \leq \mathrm{AVar}\big[\hatJ_\mathrm{x}(\pie;\D)\big],\quad  \forall \mathrm{x}\in\big\{\ips,\bips,\wips\big\}
\end{align*}
where $\mathrm{AVar}[\hatJ]$ denotes the variance of the limiting distribution of $\sqrt{n'}\,(\hatJ-J(\pie))$. This resolves the dilemma between balanced and weighted IPS noted by~\citet{agarwal2017effective}: asymptotically, $\hatJ_\oips$ is at least as efficient as both.

\paragraph{Comparison with $\hatJ_{\drbips}$.} Rewriting our infeasible optimal estimator $\hatJ(\pie,\balp^\star;\D)$~\eqref{eq:hatJ-control-variate-form}, we have
\begin{align*}
    \hatJ(\pie,\balp^\star;\D)= \frac{1}{n}\sum_{i=1}^K \sum_{(S,A,R)\in\D_i}\sbra{\frac{\pie(A\mid S)}{\piavg(A\mid S)}\rbra{R - \frac{g_{\balp^\star}}{\pie}(S,A)} + \frac{g_{\balp^\star}}{\pie}(S,\pie)},
    \numberthis\label{eq:hatJ-kallus-form}
\end{align*}
where $g_{\balp^\star}\coloneq \sum_{i\in[K]}\alp^\star_i \pii$ and $\frac{g_{\balp^\star}}{\pie}(s,\pie)\coloneq \int_\A \frac{g_{\balp^\star}(s,a)}{\pie(a\mid s)}\pie(a|s)\,d\mu_a(a)=\sum_{i\in [K]}\alp_i^\star$.

Thus, our estimator belongs to the class of estimators studied by~\citet{kallus2021optimal}:
\begin{align*}
    \hatJ_{\Gamma}(\pie,h,f;\D) = \frac{1}{n}\sum_{i=1}^K \sum_{(S,A,R)\in\D_i}\big[h(i,S,A)\,\pie(A\mid S)\big(R - f(S,A)\big) + f(S,\pie)\big],
\end{align*}
where $h$ satisfies $\sum_i(n_i/n)\,\pii(a\mid s)\,h(i,s,a)=1$, which makes $\hatJ_\Gamma$ unbiased. The optimal variance is achieved by $h(i,s,a)=1/\piavg(a\mid s)$ for all $i\in [K]$ and $f(s,a)=q(s,a)$, with $q(s,a)$ denoting the conditional mean reward. This holds even though our weights $w_i$ may depend on $(s,a,r)$ while $h$ depends only on $(i,s,a)$: the representation~\eqref{eq:hatJ-kallus-form} corresponds to $h=1/\piavg$ and $f=g_{\balp^\star}/\pie$.

The variance gap between our infeasible estimator $\hatJ(\pie,\bw^\star;\D)$, which is optimal within $\wipsclass$, and the infeasible optimal estimator $\hatJ^\star_{\Gamma}(\pie;\D)\coloneq\hatJ_{\Gamma}(\pie,1/\piavg,q;\D)$ in this class is characterized in terms of the residual from the orthogonal projection of $q\,\pie$ onto the finite-dimensional space spanned by the logging policies, $\mathrm{span}\big(\{\pii\}_{i\in[K]}\big)$. The precise statement is given below, with the proof deferred to Appendix~\ref{proof:thm:variance-gap}.
\begin{restatable}{theorem}{ThmVarianceGap}
    The variance gap between $\hatJ(\pie,\balp^\star;\D)$ and the semiparametrically efficient oracle estimator $\hatJ^\star_{\Gamma}(\pie;\D)\coloneq\hatJ_{\Gamma}(\pie,1/\piavg,q;\D)$ is given by
    \begin{align*}
        \var\big[\hatJ(\pie,\balp^\star;\D)\big] - \var\big[\hatJ^\star_{\Gamma}(\pie;\D)\big]
        = \frac{1}{n}\rbra{\inf_{\balp\in\rr^K}\big\|q\,\pie - g_{\balp}\big\|^2_{\mulog} - \var_{p_s}\big[q(S,\pie)\big]}\geq 0,
    \end{align*}
    where $\|\cdot\|_{\mulog}$ is the norm induced by the inner product $\innp{\cdot,\cdot}_{\mulog}$, with $d\mulog \coloneq p_s/\piavg\,d\mu_{s,a}$. Moreover, equality holds whenever $q\,\pie\in \mathrm{span}\big(\{\pii\}_{i\in[K]}\big)$.
    \label{thm:variance-gap}
\end{restatable}
\cite{kallus2021optimal} showed that, for any consistent estimator $\hatq$ of the conditional mean reward $q$, under their regularity conditions, $\hatJ_\drbips$~\eqref{eq:dr-bIPS} achieves the optimal asymptotic variance $n\var\big[\hatJ^\star_{\Gamma}(\pie;\D)\big]$.

Combining this result with Theorem~\ref{thm:optimalIPSasymptotics}, we obtain
\begin{align*}
    \mathrm{AVar}\big[\hatJ_\drbips(\pie;\D)\big] \leq \mathrm{AVar}\big[\hatJ_\oips(\pie;\D)\big].
\end{align*}
We close this gap in Appendix~\ref{app:dr} with a doubly robust extension that uses the same reward model: it applies the control-variate form~\eqref{eq:hatJ-control-variate-form} to the residual $R-\hatq(S,A)$ and adds the direct-method term $\hatq(S,\pie)$. We show that, with known logging policies, $\hatJ_\droips$ has asymptotic variance no larger than that of $\hatJ_\drbips$. Moreover, suppose that both estimators use the same misspecified reward model, $\hatq\to q'\neq q$. If $(q-q')\pie\in\mathrm{span}\{\pii\}_{i=1}^K$, then $\hatJ_\droips$ still attains the oracle asymptotic variance $n\var\big[\hatJ_\Gamma^\star(\pie;\D)\big]$, whereas $\hatJ_\drbips$ in general does not. The reason is that $\hatJ_\droips$ projects $(q-q')\pie$ onto $\mathrm{span}\{\pii\}_{i=1}^K$ in $L^2(\mulog)$ and effectively uses the corrected reward model $q'+g^{\Delta}/\pie$, where $g^{\Delta}$ denotes this projection; only the component of $(q-q')\pie$ orthogonal to the span contributes excess variance.

\section{Numerical Experiments}\label{sec:experiment}
\subsection{Benchmark Datasets and Experiment Setup}
\begin{wraptable}{r}{0.6\textwidth}
    \centering
    \vspace{-4mm}
    \caption{Dataset Statistics}
    \label{tab:dataset}
    \resizebox{\linewidth}{!}{
        \begin{tabular}{lcccc}
            \toprule
            \textbf{Dataset Name} & \textbf{OptDigits} & \textbf{SatImage} & \textbf{PenDigits} & \textbf{Letter} \\
            \#Classes ($l$) & 10 & 6 & 10 & 26 \\
            \#Samples & 5620 & 6435 & 10992 & 20000 \\
            \bottomrule
        \end{tabular}
    }
    \vspace{-4mm}
\end{wraptable} 
Following prior multi-logger OPE protocols~\citep{agarwal2017effective,kallus2021optimal}, we evaluate on four UCI datasets (Table~\ref{tab:dataset}): \textit{OptDigits}~\citep{misc_optical_recognition_of_handwritten_digits_80}, \textit{SatImage}~\citep{misc_statlog_(landsat_satellite)_146}, \textit{PenDigits}~\citep{misc_pen-based_recognition_of_handwritten_digits_81}, and \textit{Letter}~\citep{misc_letter_recognition_59}. Each classification dataset is converted into a contextual bandit dataset by treating labels as actions and assigning reward $1$ for a correct action and $0$ otherwise.

\paragraph{Mapping Classification Problems to Contextual Bandits.}
In deterministic-label multi-class classification, samples take the form $(S, y(S))$, where $S \sim p_s$ and $y(\cdot)$ is the ground-truth labeling function. We view this as a contextual bandit problem by treating the feature distribution $p_s$ as the context distribution and defining the deterministic reward as $r(s,a) \coloneq \ind[y(s)=a]$.

To construct the evaluation policy, we train a multinomial logistic classifier on 30\% of each dataset and define the deterministic policy $\pi_\mathrm{det}(a\mid s)\coloneq\ind[a=\hat{y}(s)]$. We set the evaluation policy to $\pie=\pi_\mathrm{det}$. Because $\pie$ is deterministic, the weights~\eqref{eq:w-star} are undefined where $\pie(a\mid s)=0$, so we compute $\hatJ_\oips$ through the control-variate form~\eqref{eq:hatJ-control-variate-form}, which does not divide by $\pie$ (Remark~\ref{rem:deterministic}). The remaining 70\% of the data is used as an evaluation pool. On this pool, the empirical accuracy of $\pi_\mathrm{det}$ is treated as the ground-truth policy value $J(\pie)$.

We generate logged bandit data from the evaluation pool using two logging policies,
\begin{align*}
    \pi_1 = 0.95\cdot\pi_\mathrm{det} + 0.05\cdot\pi_u,\qquad
    \pi_2 = 0.05\cdot\pi_\mathrm{det} + 0.95\cdot\pi_u,
\end{align*}
where $\pi_u$ is the policy that chooses an action uniformly at random for any context. Thus, $\pi_1$ is well aligned with $\pie$, whereas $\pi_2$ is much less aligned with $\pie$. For each context in the evaluation pool, we assign it to either $\D_1$ or $\D_2$, sample an action from the corresponding logging policy, and observe the deterministic reward $r(s,a)$. We vary the stratum-size ratio over $n_1/n_2\in\{0.1, 0.25, 0.5, 1, 2, 4, 10\}$. Smaller ratios make the logged dataset more dominated by $\pi_2$. Here the two loggers explore uniformly with probabilities $0.05$ and $0.95$ (the weights on $\pi_u$ above). Appendix~\ref{app:mixture_sensitivity} uses $(0.20,0.80)$ and $(0.35,0.65)$, and Appendix~\ref{app:multi_logger} uses $(0.05,0.50,0.95)$ for $K=3$ and $(0.05,0.15,0.30,0.70,0.95)$ for $K=5$.

We benchmark each estimator $\hatJ$ against the ground-truth value $J(\pie)$ using relative root mean squared error. For $M$ independent trials with ground-truth values $J_m$ and $\bar J=M^{-1}\sum_m J_m$, this metric is
\begin{align*}
    \textup{Relative-RMSE}(\hatJ) \coloneq \frac{1}{\bar J} \sqrt{ \frac{1}{M}\sum_{m=1}^M \rbra{J_m - \hatJ_m}^2 },
    \numberthis\label{eq:RelativeRMSE}
\end{align*}
where $\hatJ_m$ denotes the estimated policy value in the $m$-th trial. To separate bias from variance, we also report the squared bias $\big(M^{-1}\sum_{m=1}^M(\hatJ_m-J_m)\big)^2$ and the empirical standard deviation of $\hatJ_1,\dots,\hatJ_M$ (Appendix~\ref{app:experiment-results}). We use $M=200$ trials and report $95\%$ paired percentile-bootstrap intervals. Each of the $B=2{,}000$ bootstrap resamples draws $M$ trial indices with replacement and recomputes Relative-RMSE for every estimator from the same indices and the matching ground-truth values $J_m$; the interval endpoints are the $2.5$th and $97.5$th percentiles. Because each trial redraws the data split, the logged actions, and the cross-fitting folds, the intervals reflect all three sources of variation. To compare two estimators, we apply the same procedure to the difference of their Relative-RMSEs computed on the same resampled trials.

\paragraph{Estimators Considered.}
We compare $\hatJ_\ips$ (Eq.~\ref{eq:IPS}), $\hatJ_\bips$ (Eq.~\ref{eq:bIPS}), $\hatJ_\wips$ (Eq.~\ref{eq:wIPS}), our $\hatJ_\oips$ (\S\ref{sec:optimalIPS}), $\hatJ_\drbips$~\citep{kallus2021optimal}, and our $\hatJ_\droips$ (Appendix~\ref{app:dr}). The two DR estimators use the same reward model $\hatq$, a logistic regression fitted with two-fold cross-fitting, and the same folds: for both, every reward, propensity, and weight quantity is fitted on the training fold and evaluated on the same held-out observations. In all experiments, we compute $\hat\balp^{(z)}$ with NumPy's Moore--Penrose pseudoinverse (\texttt{numpy.linalg.pinv}, default relative cutoff $10^{-15}$). In the designs with $K\geq3$ loggers (Appendices~\ref{app:multi_logger} and~\ref{app:complexity}), every logger is a mixture of the same two policies, so $\bT$ has rank two and the pseudoinverse returns the minimum-norm least-squares solution whenever $\hat\bT^{(z)}$ is singular. Appendix~\ref{app:experiment-results} details estimation of unknown logging policies.

The experimental results are reported and discussed in Appendix~\ref{app:experiment-results}. In summary, $\hatJ_\oips$ generally improves over existing multi-logger IPS baselines across benchmark datasets and logger mixtures, and the DR extension $\hatJ_{\droips}$ achieves lower Relative-RMSE than $\hatJ_{\drbips}$ in many finite-sample settings. In a synthetic experiment with $n=10^6$ samples (Appendix~\ref{app:large-n}), $\hatJ_\oips$ attains a Relative-RMSE of $1.99\times10^{-4}$, compared with $5.44\times10^{-4}$ for $\hatJ_\wips$. Code to reproduce the experiments is included in the supplementary material and will be publicly released.

\section{Conclusion}\label{sec:conclusion}
We studied off-policy evaluation with multiple logging policies, where existing IPS estimators can have highly instance-dependent performance. We addressed this by deriving $\hatJ_\oips$, an asymptotically variance-minimizing estimator within the generalized weighted-IPS class $\wipsclass$, which contains previously considered IPS estimators. The optimal weights have a closed-form characterization through the linear system $\bT\balp=\bc$, whose entries can be estimated from logged data, leading to a practical cross-fitted estimator. This gives a principled way to choose sample-dependent IPS weights rather than selecting among fixed rules such as naive IPS, balanced IPS, and weighted IPS.

We also related $\hatJ_\oips$ to the semiparametric efficiency framework of~\citet{kallus2021optimal}. The linear system $\bT\balp=\bc$ has a projection interpretation: it chooses the best control variate representable by the logging policies. When this space is rich enough, $\hatJ_\oips$ asymptotically recovers the efficient oracle. Otherwise, the projection residual characterizes the variance gap. We further extended the estimator to a doubly robust version, $\hatJ_\droips$, which matches the asymptotic variance of the efficient DR baseline when the reward model is correctly specified and can be more robust under certain structured misspecifications.

Empirically, $\hatJ_\oips$ improves over existing multi-logger IPS baselines across benchmark datasets and logger mixtures, and $\hatJ_\droips$ often improves over the DR balanced IPS baseline in finite samples. The main limitation is that our optimality result is restricted to $\wipsclass$ and our asymptotic guarantees assume known logging policies; with estimated propensities, the limiting variance generally changes (Remark~\ref{rem:estimated}). We also do not analyze finite-sample behavior or the stability of the estimated weights, the method relies on sufficient overlap, and our experiments cover UCI benchmarks and one synthetic environment. Future work includes finite-sample analysis, richer basis classes beyond logging policies, and extensions to sequential settings.

\newpage
\begin{ack}
This research was supported by the Deep-Tech TIPS Program (Grant No. RS-2024-0046898) funded by the Ministry of SMEs and Startups, Republic of Korea.
\end{ack}
\bibliography{ref}
\bibliographystyle{abbrvnat}

\newpage
\appendix
\makeatletter
\setlength{\@fptop}{0pt}
\setlength{\@fpsep}{\floatsep}
\setlength{\@fpbot}{0pt plus 1fil}
\makeatother

\section{Main Experimental Results and Discussion}\label{app:experiment-results}
\suppressfloats[t]

\begin{figure*}[ht]
    \centering
    \includegraphics[width=\textwidth]{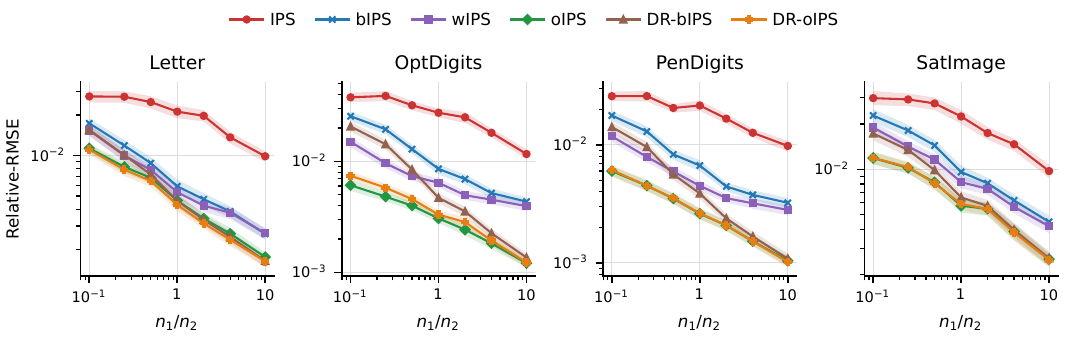}
    \caption{Relative-RMSEs~\eqref{eq:RelativeRMSE}, computed over $M = 200$ independent experiments, across datasets and estimators when the logging policies are \textbf{known}. Our proposed estimator $\hatJ_\oips$ is implemented with $Z=5$-fold cross-fitting. The $x$-axis denotes the stratum-size ratio ($n_1 / n_2$) on a logarithmic scale, while the $y$-axis shows Relative-RMSE, also on a logarithmic scale. Lower values indicate better performance. Shaded bands show $95\%$ paired percentile-bootstrap intervals (\S\ref{sec:experiment}).}
    \label{fig:results-known}
\end{figure*}

\begin{figure*}[t]
    \centering
    \includegraphics[width=\textwidth]{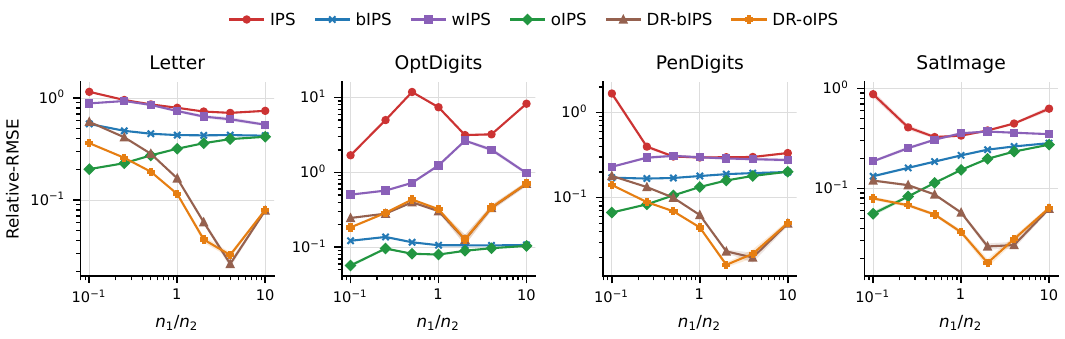}
    \caption{Relative-RMSEs~\eqref{eq:RelativeRMSE}, computed over $M = 200$ independent experiments, across datasets and estimators when the logging policies are \textbf{estimated}. Shaded bands show $95\%$ paired percentile-bootstrap intervals (\S\ref{sec:experiment}). Where bands are not visible, they are narrower than the markers.}
    \label{fig:results-estimated}
\end{figure*}

This appendix reports the full experimental results for the setup described in \S\ref{sec:experiment}. Figures~\ref{fig:results-known} and~\ref{fig:results-estimated} show Relative-RMSE on \textit{Letter}, \textit{OptDigits}, \textit{PenDigits}, and \textit{SatImage} across the stratum-size ratios $n_1/n_2\in\{0.1, 0.25, 0.5, 1, 2, 4, 10\}$. Each point is computed from $M=200$ independent trials, and our estimator $\hatJ_\oips$ uses $Z=5$ folds.

\paragraph{When Logging Policies are Unknown.}
For non-DR estimators with unknown logging policies, we estimate each logging policy $\hatpi_i$ separately using 2-fold cross-fitted multinomial logistic regression within each logger. The out-of-fold models are used to estimate the observed-action propensities $\hatpi_i(A_{ij}\mid S_{ij})$ for samples in stratum $i$. We also fit a model on all samples from stratum $i$ and use it to evaluate $\hatpi_i(A_{kj}\mid S_{kj})$ for samples from all strata, which is needed when an estimator requires the propensity of a sample under a different logging policy.

For numerical stability, we blend each predicted action distribution with a 1\% uniform distribution and clip all propensities below $10^{-3}$. For degenerate strata, such as strata with too few samples or only one observed action class, we instead use the uniform policy. The resulting estimated logging policies are then plugged into IPS-style estimators.

\paragraph{Known Logging Policies.}
When logging policies are known (Figure~\ref{fig:results-known}), $\hatJ_\oips$ consistently improves over non-DR IPS baselines across datasets and stratum-size ratios. In several regimes, it is also competitive with or better than $\hatJ_\drbips$. Adding a reward model ($\hatJ_\droips$) improves only slightly on $\hatJ_\oips$, which suggests that little residual structure remains after the control-variate correction. These results support the practical value of the control-variate form~\eqref{eq:hatJ-control-variate-form}, which uses the logger-basis structure to reduce variance without requiring a conditional reward estimator. By contrast, $\hatJ_\ips$ and $\hatJ_\bips$ often show larger errors, especially at small $n_1/n_2$, where the logged data are dominated by the less aligned logger $\pi_2$.

\paragraph{Estimated Logging Policies.}
When logging policies are estimated (Figure~\ref{fig:results-estimated}), the same qualitative trend persists: $\hatJ_\oips$ remains strong and generally outperforms non-DR baselines ($\hatJ_\ips$, $\hatJ_\bips$, $\hatJ_\wips$) across datasets and most stratum-size ratios. This is consistent with sample-dependent, cross-fitted weighting through $\hat{\bT}$ and $\hat{\bc}$, which stabilizes stratum balancing under propensity-estimation noise. In contrast, $\hatJ_{\wips}$ can underperform $\hatJ_{\bips}$ because its adaptive mixture weights depend on plug-in divergence estimates that are themselves sensitive to estimation noise.

Relative to the DR baseline $\hatJ_{\drbips}$, $\hatJ_{\oips}$ is not uniformly better when logging policies are estimated; this regime is not covered by Theorem~\ref{thm:optimalIPSasymptotics} (Remark~\ref{rem:estimated}). However, the natural DR extension $\hatJ_{\droips}$ achieves lower Relative-RMSE than $\hatJ_{\drbips}$ in many settings, particularly at low stratum-size ratios where the data are dominated by the less aligned logger. Subtracting the reward model first shrinks the inverse-propensity-weighted term $\pie(R-\hatq)/\piavg$, and $\hatJ_\droips$ then applies the control-variate correction to this smaller residual, which can improve accuracy in finite samples.

\paragraph{Summary.}
Overall, $\hatJ_{\oips}$ is a strong practical estimator for multi-logger OPE, especially against IPS baselines, and its DR extension $\hatJ_{\droips}$ is often strongest when propensities are estimated. Appendices~\ref{app:multi_logger} and~\ref{app:mixture_sensitivity} show that these trends persist with more than two loggers and under alternative mixture settings. Across all settings, $\hatJ_\oips$ has a lower Relative-RMSE than each non-DR baseline in $276$ of $276$ configuration--baseline comparisons with known propensities and in $262$ of $276$ with estimated propensities. $\hatJ_\droips$ has a lower Relative-RMSE than $\hatJ_\drbips$ in $92$ of $92$ and $66$ of $92$ configurations, respectively, and the paired $95\%$ interval for the difference lies below zero in $89$ and $64$ of them.

Table~\ref{tab:bias-sd} separates the error of each estimator into squared bias and empirical standard deviation at three representative stratum-size ratios. The squared bias is at most $1.5\%$ of the variance in every cell, so the differences in Relative-RMSE come mainly from variance, and the smallest standard deviation is attained by $\hatJ_\oips$ or $\hatJ_\droips$ in all twelve dataset--ratio pairs.

\begin{table}[t]
\centering
\caption{Squared bias (top) and empirical standard deviation (bottom) of each estimator over $M=200$ trials at three representative stratum-size ratios, with known logging policies ($\eta=0.95$, $K=2$).}
\label{tab:bias-sd}
\small
\setlength{\tabcolsep}{4pt}
\begin{tabular}{llcccccc}
\toprule
$n_1/n_2$ & Dataset & IPS & bIPS & wIPS & oIPS & DR-bIPS & DR-oIPS \\
\midrule
$10^{-1}$ & Letter & \shortstack{1.365e-6\\0.021194} & \shortstack{2.426e-7\\0.013643} & \shortstack{4.199e-9\\0.012154} & \shortstack{2.064e-7\\0.009162} & \shortstack{9.958e-8\\0.012274} & \shortstack{2.174e-7\\0.008961} \\[3pt]
 & OptDigits & \shortstack{9.845e-6\\0.036553} & \shortstack{6.086e-6\\0.025019} & \shortstack{1.955e-6\\0.015421} & \shortstack{6.030e-8\\0.007236} & \shortstack{3.771e-6\\0.020381} & \shortstack{1.661e-8\\0.008195} \\[3pt]
 & PenDigits & \shortstack{3.306e-6\\0.024435} & \shortstack{1.867e-6\\0.016799} & \shortstack{4.806e-7\\0.011213} & \shortstack{1.850e-7\\0.006118} & \shortstack{1.739e-6\\0.013490} & \shortstack{3.690e-7\\0.006194} \\[3pt]
 & SatImage & \shortstack{1.315e-7\\0.023837} & \shortstack{1.240e-7\\0.018533} & \shortstack{1.246e-7\\0.015773} & \shortstack{9.588e-8\\0.010424} & \shortstack{2.831e-7\\0.014409} & \shortstack{1.638e-7\\0.010491} \\[3pt]
\midrule
$1$ & Letter & \shortstack{2.891e-9\\0.016254} & \shortstack{1.825e-8\\0.005653} & \shortstack{2.700e-8\\0.005392} & \shortstack{6.420e-9\\0.004901} & \shortstack{2.165e-8\\0.004846} & \shortstack{1.414e-8\\0.004680} \\[3pt]
 & OptDigits & \shortstack{3.120e-6\\0.026398} & \shortstack{5.313e-8\\0.009119} & \shortstack{7.139e-8\\0.007522} & \shortstack{5.952e-9\\0.004654} & \shortstack{1.070e-9\\0.005835} & \shortstack{2.516e-8\\0.004749} \\[3pt]
 & PenDigits & \shortstack{6.577e-7\\0.020048} & \shortstack{2.949e-8\\0.006382} & \shortstack{5.040e-10\\0.004676} & \shortstack{1.553e-8\\0.003435} & \shortstack{2.238e-10\\0.004076} & \shortstack{2.223e-8\\0.003374} \\[3pt]
 & SatImage & \shortstack{8.208e-8\\0.018483} & \shortstack{9.300e-8\\0.009005} & \shortstack{9.909e-8\\0.008035} & \shortstack{4.448e-10\\0.006581} & \shortstack{7.469e-9\\0.007054} & \shortstack{9.303e-10\\0.006691} \\[3pt]
\midrule
$10$ & Letter & \shortstack{8.981e-7\\0.007988} & \shortstack{1.555e-10\\0.004009} & \shortstack{7.371e-9\\0.004053} & \shortstack{2.392e-10\\0.003561} & \shortstack{8.685e-10\\0.003556} & \shortstack{6.415e-10\\0.003532} \\[3pt]
 & OptDigits & \shortstack{1.723e-9\\0.011590} & \shortstack{8.947e-8\\0.005328} & \shortstack{1.261e-7\\0.005066} & \shortstack{1.026e-9\\0.003627} & \shortstack{3.153e-9\\0.003636} & \shortstack{7.132e-10\\0.003605} \\[3pt]
 & PenDigits & \shortstack{5.843e-7\\0.009378} & \shortstack{1.267e-8\\0.003771} & \shortstack{9.083e-13\\0.003493} & \shortstack{5.156e-10\\0.002592} & \shortstack{1.545e-9\\0.002580} & \shortstack{4.439e-10\\0.002573} \\[3pt]
 & SatImage & \shortstack{3.424e-7\\0.008730} & \shortstack{5.540e-8\\0.005562} & \shortstack{2.232e-8\\0.005446} & \shortstack{1.047e-8\\0.005163} & \shortstack{8.144e-9\\0.005223} & \shortstack{5.826e-9\\0.005246} \\[3pt]
\bottomrule
\end{tabular}
\end{table}

\section{Doubly Robust Extension of $\hatJ_\oips$}
\label{app:dr}
In this section, we present a doubly robust (DR) extension of our variationally optimal estimator $\hatJ_\oips$. Given any conditional reward estimator $\hatq:\calS\times\A\to\rr$, we show how to incorporate $\hatq$ to obtain a DR version of $\hatJ_\oips$, denoted by $\hatJ_{\droips}$. We then compare $\hatJ_{\droips}$ both theoretically and empirically with the semiparametrically efficient estimator $\hatJ_{\drbips}$.

\subsection{Generalized Weighted DR-IPS Estimator Class $\drwipsclass$}
We define the generalized weighted DR-IPS estimator class $\drwipsclass$ as follows. Let $\hatq:\calS\times\A\to\rr$ be an estimator of the conditional mean reward $q(s,a)\coloneq\expec[R\,|\, S=s,A=a]$, and define
\begin{align*}
    \hatJ_\dm(\pie,\hatq;\D) \coloneq\frac{1}{n}\sum_{i=1}^K\sum_{(S,A,R)\in\D_i} \int_\A \hatq(S, a)\,\pie(a\mid S)\,d\mu_a(a)
\end{align*}
as the direct value estimator induced by $\hatq$.

Then, the estimator class $\drwipsclass$ consists of estimators of the following form:
\begin{align*}
    \hatJ(\pie, \bw, \hatq; \D) \coloneq \hatJ_\dm(\pie,\hatq;\D) + \sum_{i=1}^K \frac{1}{n_i} \sum_{(S,A,R)\in\D_i} w_i(S,A,R)\cdot\frac{\pie(A\mid S)}{\pi_i(A\mid S)}.
    \numberthis\label{eq:gwDR_def}
\end{align*}
where $\Delta(s,a,r)\coloneq r-\hatq(s,a)$ is the residual. The estimator is unbiased when $\hatq$ is a fixed function, or conditionally on the training data when $\hatq$ is fitted on data independent of $\D$, provided the weights satisfy the following constraints:
\begin{enumerate}[leftmargin=*,label=(\roman*)]
    \item \textbf{Normalization:} $\sum_{i=1}^K w_i(s,a,r) = \Delta(s,a,r)$,
    \item \textbf{Dominance:} $\pi_i(a|s) = 0 \implies w_i(s,a,r) = 0$.
\end{enumerate}

Intuitively, this form follows from the decomposition
\begin{align*}
    \expec_{\mupie}[R] = \expec_{\mupie}[R-\hatq]+\expec_{\mupie}[\hatq].
\end{align*}
That is, we apply weighted IPS to the residual term and use the direct method for the second term.

\subsection{Feasible Extension: $\hatJ_\droips$}
We seek to minimize the variance of~\eqref{eq:gwDR_def}. The variational approach used for $\hatJ_\oips$ applies directly here, with the target variable $r$ replaced by the residual $\Delta(s,a,r)\coloneq r-\hatq(s,a)$.

Following the same cross-fitting procedure described in \S\ref{sec:optimalIPS} for constructing the feasible estimator $\hatJ_\oips$, we obtain a feasible doubly robust extension, denoted $\hatJ_\droips$. For each fold $z$, let $\hatq^{(z)}$ denote the conditional reward estimator trained on $\D^{(-z)}$. On the held-out fold $\D^{(z)}$, define the fold-specific direct-method term by
\begin{align*}
    \hatJ_\dm\rbra{\pie,\hatq^{(z)};\D^{(z)}} 
    \coloneq
    \frac{1}{n^{(z)}}\sum_{i=1}^K\sum_{(S,A,R)\in\D_i^{(z)}} \int_\A \hatq^{(z)}(S, a)\,\pie(a\mid S)\,d\mu_a(a).
\end{align*}
The residual IPS component is obtained by applying the control-variate form~\eqref{eq:hatJ-control-variate-form} to the residuals $R-\hatq^{(z)}(S,A)$. Specifically, let
\begingroup\small
\begin{align*}
    \hatJ_\Delta\!\rbra{\pie,\hat{\balp}^{(z)};\D^{(z)}}&\\
    &\hspace{-1in}\coloneq
    \frac{1}{n^{(z)}}\sum_{i\in[K]}\sum_{(S,A,R)\in\D_i^{(z)}}
    \sbra{
        \frac{\big(R-\hatq^{(z)}(S,A)\big)\pie(A\mid S)}{\piavg^{(z)}(A\mid S)}
        -\frac{\sum_{k\in[K]}\hat{\alp}^{(z)}_k\pik(A\mid S)}{\piavg^{(z)}(A\mid S)}
        +\sum_{k\in[K]}\hat{\alp}^{(z)}_k
    },
\end{align*}
\endgroup
where
\begin{align*}
    \piavg^{(z)}\coloneq \frac{1}{n^{(z)}}\sum_{i\in [K]} n_i^{(z)} \pii.
\end{align*}
The feasible $\droips$ estimator is then
\begin{align*}
\boxed{
    \hatJ_\droips(\pie;\D)\coloneq\frac{1}{n}\sum_{z=1}^Z n^{(z)}\sbra{\hatJ_\dm\rbra{\pie,\hatq^{(z)};\D^{(z)}}+\hatJ_\Delta\!\rbra{\pie,\hat{\balp}^{(z)};\D^{(z)}}}
    \numberthis\label{eq:stable_dr_score}
}
\end{align*}

\paragraph{Shared outer-fold procedure.}
Both DR estimators split each stratum into the same two folds ($Z=2$). For each fold $z$:
\begin{enumerate}[leftmargin=*,label=\arabic*.]
    \item On the training data $\D^{(-z)}$, fit the reward model $\hatq^{(z)}$, the logging-policy models when the propensities are unknown, and the moments $\hat\bT^{(z)}$ and $\hat\bc^{(z)}$ defined below, and set $\hat\balp^{(z)}=(\hat\bT^{(z)})^\dagger\hat\bc^{(z)}$.
    \item Evaluate both estimators on the held-out data $\D^{(z)}$, using $\hat\balp^{(z)}$ for $\hatJ_\droips$ and $\hat\balp^{(z)}=0$ for $\hatJ_\drbips$.
\end{enumerate}
The fold estimates are combined as in~\eqref{eq:stable_dr_score}. With $\hat\balp^{(z)}=0$, this yields the cross-fitted form of~\eqref{eq:dr-bIPS} with $\hatq^{(z)}$ and $\piavg^{(z)}$.

\paragraph{Estimation of Coefficients.} As in $\hatJ_\oips$, the parameters $\hat{\balp}^{(z)}$ in~\eqref{eq:stable_dr_score} are obtained from $\bT\balp=\bc$ via cross-fitting. Specifically, $\hat{\balp}^{(z)}=(\hat{\bT}^{(z)})^\dagger\hat{\bc}^{(z)}$, where $\hat{\bT}^{(z)}$ and $\hat{\bc}^{(z)}$ are estimated on $\D^{(-z)}$ as follows.

\begin{enumerate}[leftmargin=*,label=(\roman*)]
    \item \textbf{Technique Matrix.} $\bT$ depends only on propensities and sample sizes, so it is identical to the IPS case:
    \begin{align*}
        \hatT_{ij}^{(z)} \coloneq \frac{1}{\big|\D^{(-z)}_i\big|} \sum_{(S,A,R) \in \D^{(-z)}_i} \frac{\pi_j(A\mid S)}{\piavg^{(-z)}(A\mid S)}.
    \end{align*}

    \item \textbf{Contribution Vector.}
    The vector $\bc$ depends on the target signal. For DR, we use residuals $R-\hatq^{(z)}(S,A)$ in place of $R$:
    \begin{align*}
        \hatc_i^{(z)} \coloneq \frac{1}{\big|\D^{(-z)}_i\big|} \sum_{(S,A,R)\in \D^{(-z)}_i}\frac{\,\big(R-\hatq^{(z)}(S,A)\big)\,\pie(A\mid S)}{\piavg^{(-z)}(A\mid S)}.
    \end{align*}
\end{enumerate}
Here, $\piavg^{(-z)}\coloneq \frac{1}{n^{(-z)}}\sum_{i\in [K]} n_i^{(-z)} \pii$, with $n^{(-z)}\coloneq \sum_{i\in[K]}n_i^{(-z)}$, as in the main text, \S\ref{subsec:feasible-ips}. When the logging policies are unknown, we replace the propensities in the above expressions by their cross-fitted estimates.

\subsection{Theoretical Comparison with the Optimal Estimator of~\cite{kallus2021optimal}}
\label{subsec:dr-theory}
In this section, we provide a theoretical comparison between $\hatJ_\droips$ and $\hatJ_\drbips$. The purpose is to explain how the residual version of our variational construction relates to the semiparametrically efficient estimator studied in~\citet{kallus2021optimal}.

The comparison has two implications. 
\begin{enumerate}[leftmargin=*,label=(\roman*)]
    \item When $\hatq$ is a consistent estimator of the conditional mean reward $q(s,a)$, both $\hatJ_\droips$ and $\hatJ_\drbips$ attain the \textbf{same} semiparametrically efficient oracle variance asymptotically.
    
    \item When $\hatq$ is \textbf{misspecified}, $\hatJ_\droips$ can remain variance-optimal in cases where the misspecification lies in the logger-basis space, whereas $\hatJ_\drbips$ need not.
\end{enumerate}
With known logging policies and under the conditions of Theorem~\ref{thm:optimalIPSasymptotics}, these conclusions follow from the following lemma, proved in Appendix~\ref{app:proof}.

\begin{restatable*}[Variance Gap]{lemma}{LemVarianceGap}
    Let $f$ be measurable and such that the summands of the estimator below have finite second moments under each $\mupii$, and let $\hatJ_{\Gamma}(\pie,f;\D)$ be the estimator defined by
    \begin{align*}
        \hatJ_{\Gamma}(\pie,f;\D)\coloneq\frac{1}{n}\sum_{i=1}^K \sum_{(S,A,R)\in\D_i}\sbra{\frac{\pie(A\mid S)}{\piavg(A\mid S)}\big(R - f(S,A)\big) + f(S,\pie)},
    \numberthis\label{eq:hatJgamma}
    \end{align*}
    where $f(s,\pie)\coloneq \int_{\A} f(s,a)\pie(a|s)\,d\mu_a(a)$.

    Let $q(s,a)\coloneq\expec[R\,|\,S=s, A=a]$ denote the conditional mean reward, and define $m_i(f)\coloneq\expec_{\mupii}[(q-f)\pie/\piavg]-\expec_{p_s}[(q-f)(S,\pie)]$. Then, the variance gap between $\hatJ_{\Gamma}(\pie,f;\D)$ and $\hatJ^\star_{\Gamma}(\pie;\D)\coloneq \hatJ_{\Gamma}(\pie,q;\D)$ satisfies
    \begin{align*}
        \var\big[\hatJ_{\Gamma}(\pie,f;\D)\big] &- \var\big[\hatJ^\star_{\Gamma}(\pie;\D)\big]\\
        &\hspace{-0.7in}= \frac{1}{n}\rbra{\var_{\mupiavg}\!\sbra{\frac{\pie(A\mid S)}{\piavg(A\mid S)}\big(q(S,A) - f(S,A)\big)} - \var_{p_s}\big[q(S,\pie) - f(S,\pie)\big]}\\
        &\hspace{-0.7in}\quad-\frac{1}{n}\sum_i\frac{n_i}{n}m_i(f)^2\geq 0.
    \end{align*}
    \label{lem:variance-gap}
\end{restatable*}
Below, the \emph{asymptotic excess variance} of an estimator is its $\mathrm{AVar}$ minus $n\var[\hatJ^\star_{\Gamma}(\pie;\D)]$, which equals $n$ times the gap in Lemma~\ref{lem:variance-gap} evaluated at the limiting $f$.

\paragraph{Consistent reward estimation.}
We first explain how $\hatJ_\droips$ fits into the $\hatJ_\Gamma$ class. For fixed $\hatq$ and $\balp$, the infeasible version of $\hatJ_\droips$ has summand
\begin{align*}
    \hatq(S,\pie)
    +
    \frac{\pie(A\mid S)}{\piavg(A\mid S)}
    \big(R-\hatq(S,A)\big)
    -
    \frac{g_{\balp}(S,A)}{\piavg(A\mid S)}
    +
    \sum_{i\in[K]}\alp_i.
\end{align*}
Since
\begin{align*}
    \sum_{i\in[K]}\alp_i
    =
    \frac{g_{\balp}}{\pie}(S,\pie),
\end{align*}
the last two terms simply replace $\hatq$ by $\hatq+g_{\balp}/\pie$ in the DR regression function. Indeed, the above summand is equal to
\begin{align*}
    \frac{\pie(A\mid S)}{\piavg(A\mid S)}
    \rbra{
        R-\rbra{\hatq(S,A)+\frac{g_{\balp}(S,A)}{\pie(A\mid S)}}
    }
    +
    \rbra{\hatq+\frac{g_{\balp}}{\pie}}(S,\pie).
\end{align*}
Thus, the infeasible $\droips$ estimator corresponds to
\begin{align*}
    \hatJ_\Gamma\!\rbra{\pie,\hatq+\frac{g_{\balp}}{\pie};\D}.
\end{align*}
The role of $\balp$ is to choose $g_{\balp}$ as the projection of $(q-\hatq)\pie$ onto $\mathrm{span}\big(\{\pii\}_{i\in[K]}\big)$. In other words, $\hatJ_\droips$ corrects $\hatq$ only in directions that can be represented by the logging policies.

If $\hatq$ is consistent for $q$, then $(q-\hatq)\pie$ converges to zero, so its projection $g_{\balp}$ also converges to zero. Therefore,
\begin{align*}
    \hatq+\frac{g_{\balp}}{\pie}\to q.
\end{align*}
Applying Lemma~\ref{lem:variance-gap} to the limiting regression function $f=q$ shows that the asymptotic excess variance of $\hatJ_\droips$ over the oracle estimator $\hatJ^\star_{\Gamma}(\pie;\D)\coloneq \hatJ_{\Gamma}(\pie,q;\D)$ vanishes. By~\citet{kallus2021optimal}, the same is true for $\hatJ_\drbips$ when $\hatq$ is consistent for $q$. Hence, in the well-specified case, $\hatJ_\droips$ and $\hatJ_\drbips$ have the same asymptotic variance, namely $n\var\big[\hatJ^\star_{\Gamma}(\pie;\D)\big]$.

\paragraph{Misspecified reward estimation.}
Now suppose that $\hatq$ converges to a misspecified limit $q'$. Then $\hatJ_\drbips$ corresponds asymptotically to
\begin{align*}
    \hatJ_\Gamma(\pie,q';\D),
\end{align*}
so Lemma~\ref{lem:variance-gap} characterizes its asymptotic excess variance over the oracle by taking $f=q'$.

By contrast, $\hatJ_\droips$ corresponds asymptotically to
\begin{align*}
    \hatJ_\Gamma\!\rbra{\pie,q'+\frac{g_{\balp^\star}}{\pie};\D},
\end{align*}
where $g_{\balp^\star}$ is the projection of $(q-q')\pie$ onto $\mathrm{span}\big(\{\pii\}_{i\in[K]}\big)$. Therefore, compared with $\hatJ_\drbips$, $\hatJ_\droips$ replaces the misspecified regression function $q'$ by the corrected regression function
\begin{align*}
    q'+\frac{g_{\balp^\star}}{\pie}.
\end{align*}
This removes the component of the misspecification that lies in the span of the logging policies. Lemma~\ref{lem:variance-gap} then implies that the asymptotic excess variance of $\hatJ_\droips$ is no larger than that of $\hatJ_\drbips$ when both use the same limiting reward model $q'$. Indeed, let $Q_f\coloneq\pie(A\mid S)\big(R-f(S,A)\big)/\piavg(A\mid S)+f(S,\pie)$ denote the summand of~\eqref{eq:hatJgamma} and $g^\star\coloneq g_{\balp^\star}$. Then $Q_{q'}=Q_{q'+g^\star/\pie}+g^\star/\piavg-\sum_k\alp^\star_k$, $m_i(q'+g^\star/\pie)=0$ for every $i$, and $\cov_{\mupiavg}\big[Q_{q'+g^\star/\pie},\,g^\star/\piavg\big]=0$, so $n\big(\var[\hatJ_\Gamma(\pie,q';\D)]-\var[\hatJ_\Gamma(\pie,q'+g^\star/\pie;\D)]\big)=\sum_i\frac{n_i}{n}\var_{\mupii}[g^\star/\piavg]\geq0$.

In particular, if
\begin{align*}
    (q-q')\pie \in \mathrm{span}\big(\{\pii\}_{i\in[K]}\big),
\end{align*}
then $g_{\balp^\star}=(q-q')\pie$, and hence
\begin{align*}
    q'+\frac{g_{\balp^\star}}{\pie}=q.
\end{align*}
Thus, $\hatJ_\droips$ attains the oracle asymptotic variance even though $\hatq$ is misspecified. In contrast, $\hatJ_\drbips$ remains tied to $q'$ and may have nonzero asymptotic excess variance. This is the sense in which $\hatJ_\droips$ is more robust to misspecification in directions spanned by the logging policies.

\subsection{Empirical Comparison with the Optimal Estimator of~\cite{kallus2021optimal}}
\begin{figure*}[ht]
    \centering
    \includegraphics[width=\textwidth]{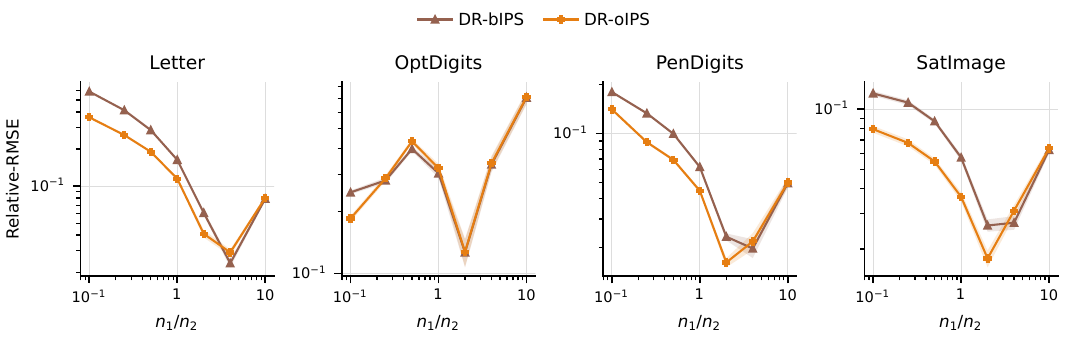}
    \caption{Relative-RMSEs, computed over $M = 200$ independent experiments, are reported across different datasets and estimators under the setting where the logging policies are \textbf{estimated}. Shaded bands show $95\%$ paired percentile-bootstrap intervals (\S\ref{sec:experiment}). Where bands are not visible, they are narrower than the markers.}
    \label{fig:results-estimated-9505-DR}
\end{figure*}

We provide an alternative, side-by-side presentation of the experimental results from \S\ref{sec:experiment}, in Figure~\ref{fig:results-estimated-9505-DR} to more directly compare $\hatJ_\droips$ with $\hatJ_\drbips$, which is known to be asymptotically optimal within the class studied by~\citet{kallus2021optimal}. We also summarize the nuisance estimation procedures used in these experiments.

\paragraph{Nuisance Estimation.}
We summarize how nuisance components (logging policies and conditional rewards) are estimated.

\begin{itemize}[leftmargin=*]
    \item \textbf{Logging policies ($\hatpi_i$):} We estimate logging policies with multinomial logistic regression and 2-fold cross-fitting. To maintain numerical stability in both the technique-matrix estimate $\hat{\bT}$ and the
    denominator of Eq.~\eqref{eq:stable_dr_score}, we clip the estimated propensities to the range $[10^{-3},1]$ after $1\%$ uniform smoothing.

    \item \textbf{Conditional rewards ($\hatq$):} We estimate $q(s,a)\coloneq\expec[r|s,a]$ via logistic regression with 2-fold cross-fitting, using the same protocol as for logging policies. Actions are represented with one-hot encodings.
\end{itemize}

\paragraph{Findings.} In the setting of Figure~\ref{fig:results-estimated-9505-DR}, $\hatJ_\droips$ has a lower Relative-RMSE than $\hatJ_\drbips$ in $16$ of $28$ configurations. Across all $184$ configurations ($92$ with known and $92$ with estimated propensities), it has a lower Relative-RMSE in $158$, and the paired $95\%$ interval for the difference lies below zero in $153$.

\section{Empirical Results with $K > 2$ Logging Policies}\label{app:multi_logger}
We extend the two-logger experiment to heterogeneous mixtures with $K \in \{3,5\}$ logging policies. As before, we train a multinomial logistic regression model on a training split to obtain a deterministic base policy $\pi_\mathrm{det}$ and set the evaluation policy to $\pie = \pi_\mathrm{det}$. The reward is deterministic: $r(s,a) \coloneq \ind\sbra{y(s)=a}$, where $y(s)$ is the label of context $s \in \rr^d$. The ground-truth policy value is therefore the classification accuracy of $\pie$ on the OPE sample.
\begin{align*}
    J(\pie)=\expec_{A\sim\pie(\cdot\mid S) , S\sim p_s}\big[\ind\sbra{y(S)=A}\big].
\end{align*}

We construct $K \in \{3,5\}$ logging policies by mixing $\pi_\mathrm{det}$ with the uniform policy $\pi_u$ (which ignores context and selects actions uniformly). For $K=5$, let $\beta_i \in \{0.05,0.15,0.30,0.70,0.95\}$ and define, for $i=1,\dots,K$,
\begin{align*}
    \pi_i(a|s) \coloneq (1-\beta_i)\cdot \pi_\mathrm{det}(a|s) + \frac{\beta_i}{|\A|}.
\end{align*}
The OPE sample is generated from an equal-weight logger mixture, with stratum weights $\tau_i = 0.2$ for $K=5$ and $\tau_i = \tfrac{1}{3}$ for $K=3$ (so $n_i/n \approx \tau_i$, up to integer rounding). We partition the OPE sample into these strata, sample $a \sim \pi_i(\cdot \mid s)$ in stratum $i$, and assign reward $r(s,a) \coloneq \ind[y(s)=a]$. For $K=3$, we use $\beta_i \in \{0.05,0.5,0.95\}$.

\begin{figure}[ht]
  \centering
  \includegraphics[width=\textwidth]{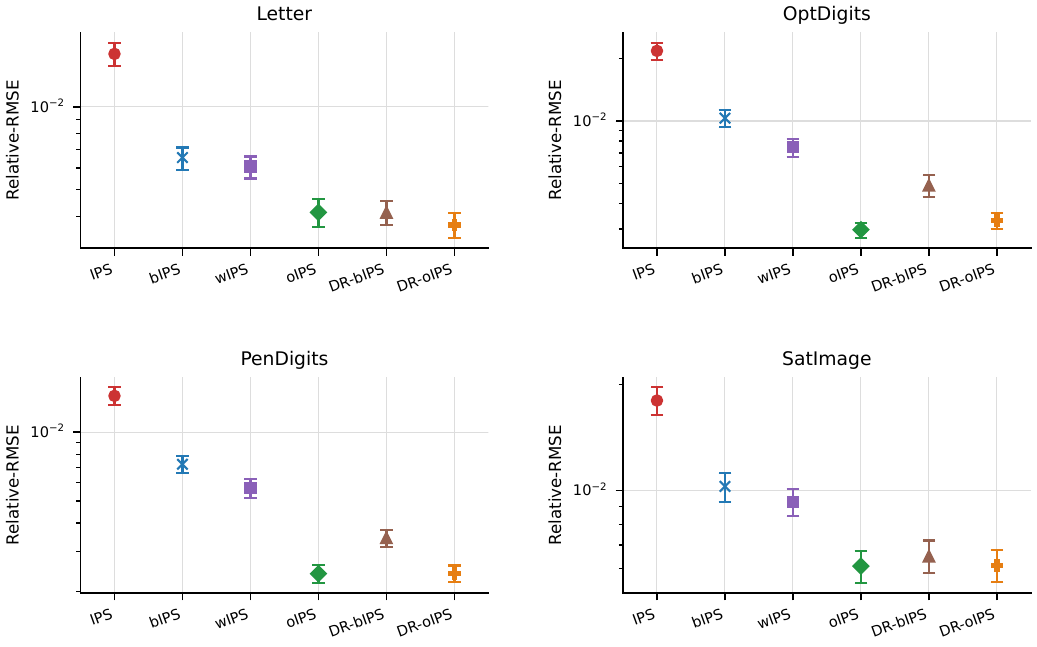}
  \caption{Rel-RMSE across five heterogeneous logging policies where they are known. Lower is better. Error bars show $95\%$ paired percentile-bootstrap intervals (\S\ref{sec:experiment}).}
  \label{fig:five-logger-known}
\end{figure}

\begin{figure}[tp]
  \centering
  \includegraphics[width=\textwidth]{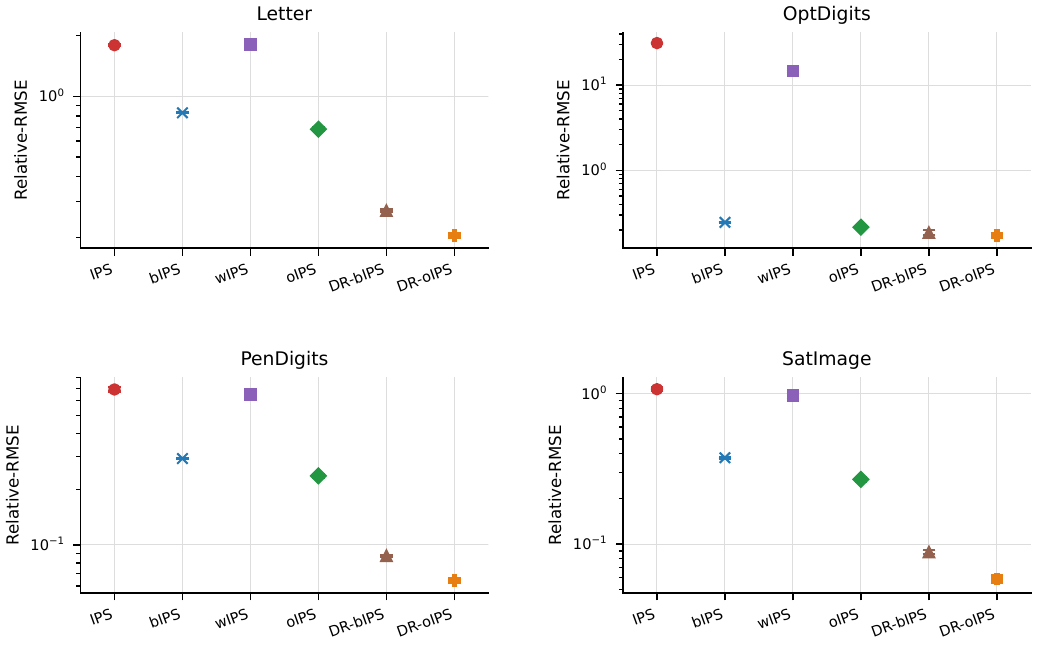}
  \caption{Rel-RMSE across five heterogeneous logging policies where they are estimated. Lower is better. Error bars show $95\%$ paired percentile-bootstrap intervals (\S\ref{sec:experiment}). Where error bars are not visible, they are shorter than the markers.}
  \label{fig:five-logger-estimated}
\end{figure}

\paragraph{Results.}
Across settings with up to five loggers (Figure~\ref{fig:five-logger-known} and Figure~\ref{fig:five-logger-estimated}), our logger-basis IPS estimator $\hatJ_{\oips}$ generally achieves strong accuracy when the logging policies are known, while its doubly robust extension $\hatJ_{\droips}$ tends to perform best when the logging policies are estimated.

\begin{figure}[tp]
  \centering
  \includegraphics[width=\textwidth]{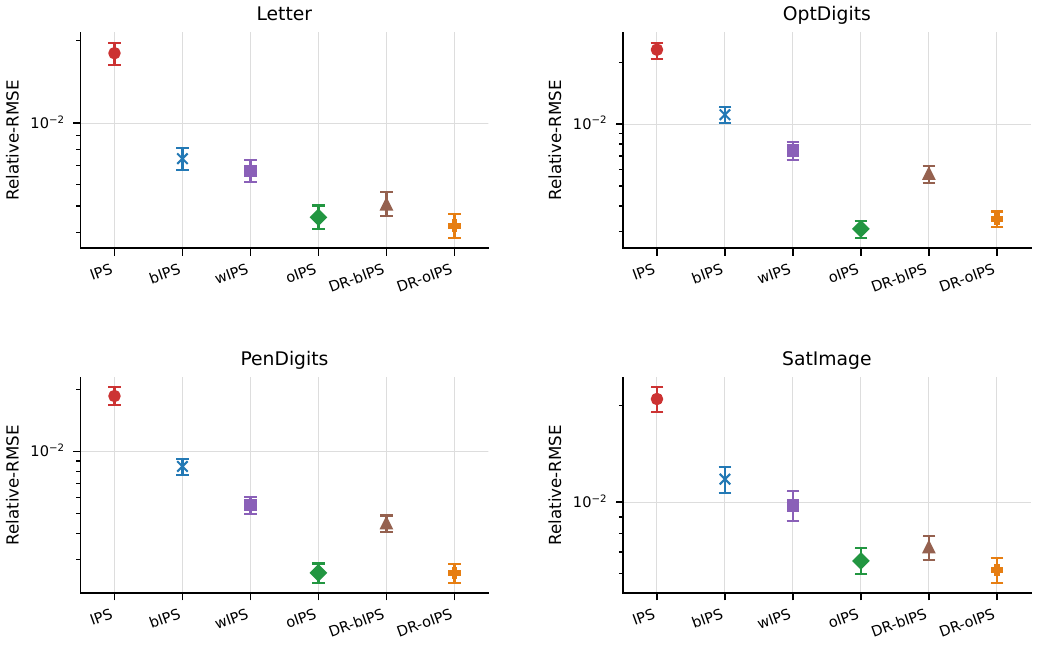}
  \caption{Rel-RMSE across three heterogeneous logging policies where they are known. Lower is better. Error bars show $95\%$ paired percentile-bootstrap intervals (\S\ref{sec:experiment}).}
  \label{fig:three-logger-known}
\end{figure}

\begin{figure}[tp]
  \centering
  \includegraphics[width=\textwidth]{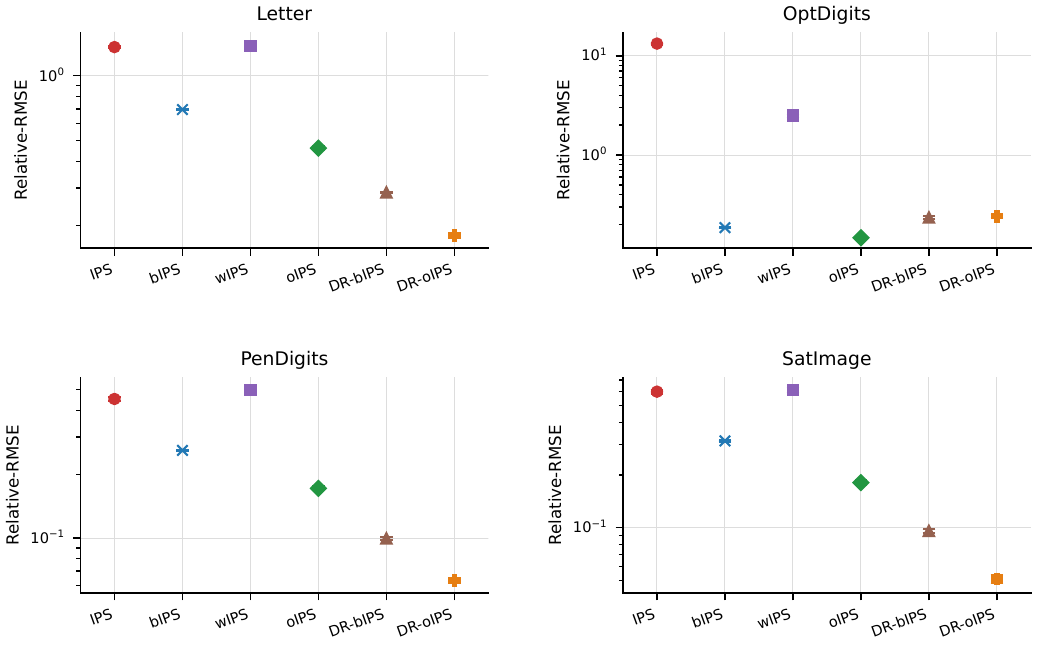}
  \caption{Rel-RMSE across three heterogeneous logging policies where they are estimated. Lower is better. Error bars show $95\%$ paired percentile-bootstrap intervals (\S\ref{sec:experiment}). Where error bars are not visible, they are shorter than the markers.}
  \label{fig:three-logger-estimated}
\end{figure}

When logging policies are estimated, $\hatJ_{\droips}$ outperforms $\hatJ_{\drbips}$ on all four datasets for $K=5$ and three of four for $K=3$. Among the non-DR estimators, $\hatJ_{\ips}$ and $\hatJ_{\wips}$ can become unstable (particularly on OptDigits). This instability appears when estimated logging propensities are near zero or noisy in specific strata, yielding extreme $1/\hatpii(a|s)$ weights and heavy-tailed variance. By comparison, the sample-dependent weighting in $\hatJ_{\oips}$ and $\hatJ_{\droips}$ better controls variance and produces substantially more stable estimates. The same qualitative pattern holds for three loggers (Figures~\ref{fig:three-logger-known} and~\ref{fig:three-logger-estimated}).

\clearpage
\section{Sensitivity to the Additive Mixture of Logging Policies}\label{app:mixture_sensitivity}

We repeat the main experiments with alternative mixture ratios between the deterministic policy $\pi_\mathrm{det}$ and the uniform policy $\pi_u$. For $\eta \in \{0.80, 0.65\}$, define
\begin{align*}
    \pi^{(\eta)}_1 \;=\; \eta\,\pi_\mathrm{det} + (1-\eta)\,\pi_u,
    \quad
    \pi^{(\eta)}_2 \;=\; (1-\eta)\,\pi_\mathrm{det} + \eta\,\pi_u.
\end{align*}
All other settings, including the evaluation policy $\pie = \pi_\mathrm{det}$, datasets, train/evaluation split ratio, stratum-size ratios $n_1/n_2$, number of repeats $M=200$, and non-DR cross-fitting folds $Z=5$, match the main-text protocol. Both DR estimators use the shared two-fold procedure of Appendix~\ref{app:dr}.

\begin{figure*}[ht]
    \centering
    \includegraphics[width=\textwidth]{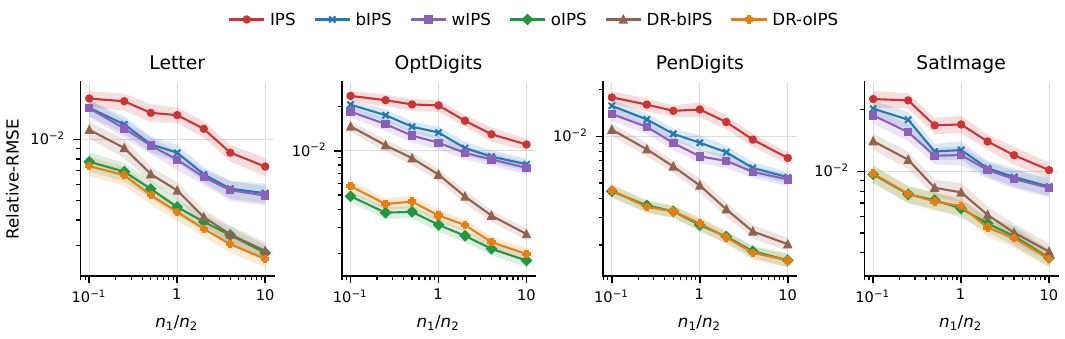}
    \caption{\textbf{Mixture $\eta=0.80$.}
    Relative-RMSEs, computed over $M = 200$ independent experiments, are reported across different datasets and estimators under the setting where the logging policies are \textbf{known}. Shaded bands show $95\%$ paired percentile-bootstrap intervals (\S\ref{sec:experiment}).}
    \label{fig:results-known-8020}
\end{figure*}

\begin{figure*}[ht]
    \centering
    \includegraphics[width=\textwidth]{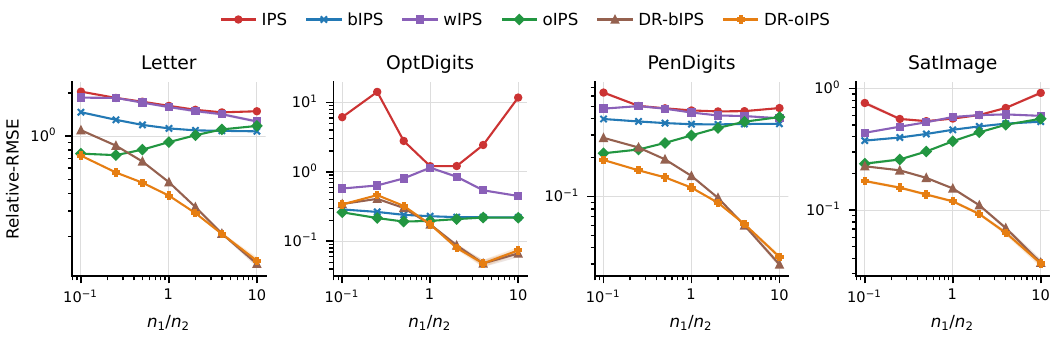}
    \caption{\textbf{Mixture $\eta=0.80$.}
    Relative-RMSEs, computed over $M = 200$ independent experiments, are reported across different datasets and estimators under the setting where the logging policies are \textbf{estimated}. Shaded bands show $95\%$ paired percentile-bootstrap intervals (\S\ref{sec:experiment}). Where bands are not visible, they are narrower than the markers.}
    \label{fig:results-estimated-8020}
\end{figure*}

\begin{figure*}[ht]
    \centering
    \includegraphics[width=\textwidth]{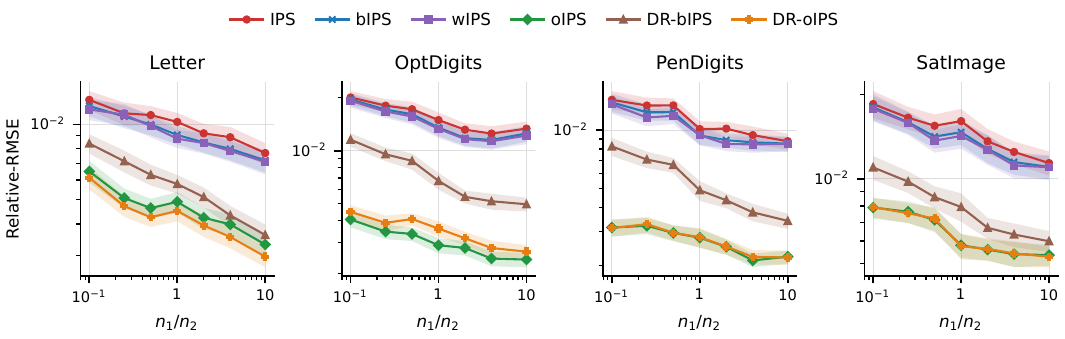}
    \caption{\textbf{Mixture $\eta=0.65$.}
    Relative-RMSEs, computed over $M = 200$ independent experiments, are reported across different datasets and estimators under the setting where the logging policies are \textbf{known}. Shaded bands show $95\%$ paired percentile-bootstrap intervals (\S\ref{sec:experiment}).}
    \label{fig:results-known-6535}
\end{figure*}

\begin{figure*}[ht]
    \centering
    \includegraphics[width=\textwidth]{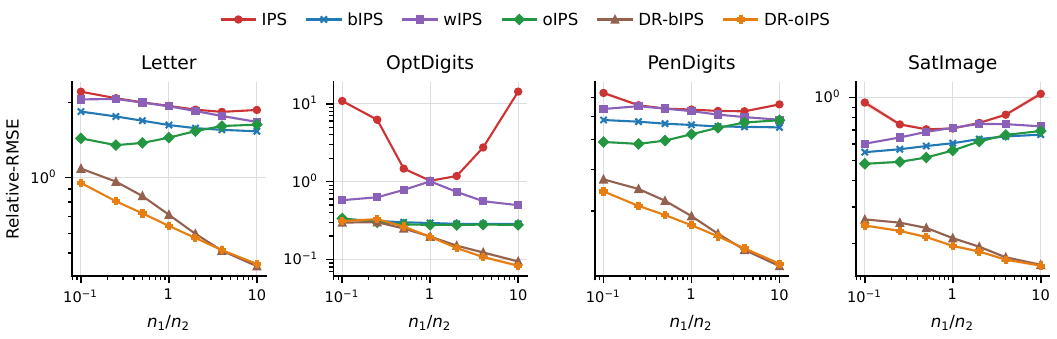}
    \caption{\textbf{Mixture $\eta=0.65$.}
    Relative-RMSEs, computed over $M = 200$ independent experiments, are reported across different datasets and estimators under the setting where the logging policies are \textbf{estimated}. Shaded bands show $95\%$ paired percentile-bootstrap intervals (\S\ref{sec:experiment}). Where bands are not visible, they are narrower than the markers.}
    \label{fig:results-estimated-6535}
\end{figure*}

\paragraph{Results.}
Across both additive mixtures ($\eta = 0.80$ and $\eta = 0.65$), the qualitative trends in Figures~\ref{fig:results-known-8020},~\ref{fig:results-estimated-8020},~\ref{fig:results-known-6535}, and~\ref{fig:results-estimated-6535} remain consistent with the main-text results. Our logger-basis IPS estimator and its DR extension ($\hatJ_{\oips}$ and $\hatJ_{\droips}$) remain among the strongest methods in known- and estimated-logging settings, respectively. Improvements are most pronounced in low-overlap regimes (small $n_1/n_2$), while gaps narrow as overlap increases. Estimated propensities increase absolute errors and can change which estimator ranks first.

\clearpage
\section{Deferred Proofs}\label{app:proof}
In this section, we present detailed proofs of the main theorems/lemmas.

\subsection{Proof of Unbiasedness of Generalized Weighted-IPS Estimators}
\Junbiased*
\begin{proof}[Proof of Theorem \ref{thm:unbiasedness}.]\label{proof:thm:unbiasedness}
Let $\omega\coloneq(s,a,r)\in\Omega$ and $\mupii$ be the probability distribution with density $p_r(r|s,a)\,\pi_i(a|s)\, p_s(s)$ with respect to $\mu$. Recall the general weighted IPS estimator,
\begin{align*}
    \hatJ(\pie,\bw;\D)&\coloneq\sum^K_{i=1} \frac{1}{n_i} \sum^{n_i}_{j=1} \frac{w_i(S_{ij},A_{ij},R_{ij})\,\pie(A_{ij}\mid S_{ij})}{\pii(A_{ij}\mid S_{ij})}\coloneq \sum^K_{i=1}\frac{1}{n_i}\sum^{n_i}_{j=1} \hatJ_{ij}(w_i;\D_i).
\end{align*}

Taking the expectation over data,
\begin{align*}
    \expec\big[\hatJ(\pie,\bw;\D)\big]&=\sum^K_{i=1}\frac{1}{n_i}\sum_{j=1}^{n_i}\expec\big[\hatJ_{ij}(w_i;\D_i)\big]\\
    &=\sum^K_{i=1}\frac{1}{n_i}\sum_{j=1}^{n_i}\int_{\supp(w_i)}w_i(\omega)\,\frac{\pie(a|s)}{\pi_i(a|s)} \,d\mupii(\omega) \tag{as $\pi_i(a|s)=0\implies w_i(s,a,r)=0$}\\
    &=\sum^K_{i=1}\frac{1}{n_i}\sum_{j=1}^{n_i}\int_{\supp(w_i)}w_i(\omega)\,\frac{\pie(a|s)}{\pi_i(a|s)}\, p_r(r|s,a)\, \pi_i(a|s)\, p_s(s)\,d\mu(\omega)\\
    &=\sum^K_{i=1}\int_{\supp(w_i)} w_i(\omega)\, p_r(r|s,a)\, \pie(a|s) \, p_s(s)\,d\mu(\omega)\\
    &=\sum^K_{i=1}\int_\Omega w_i(\omega)\,d\mupie(\omega) \\
    &=\int_\Omega\rbra{\sum^K_{i=1} w_i(\omega)}\,d\mupie(\omega)\\
    &=\int_\Omega r \, \,d\mupie(\omega)\tag{as $\sum^K_{i=1} w_i(\omega)=r$ over $\omega\in\supp(\mupie)$}\\
    &=\expec_{\mupie}[R].
\end{align*}
\end{proof}
\subsection{Proof of Theorem~\ref{thm:optimalIPS}.}\label{app:subsec:proof:thm:optimalIPS}
In this section, we prove Theorem~\ref{thm:optimalIPS}. Let $\bw^\star\coloneq (w^\star_i)_{i\in [K]}$ be defined by
\begin{align*}
    w^\star_i(s,a,r)=
    \frac{\alp^\star_i\pii(a|s)}{\pie(a|s)}
    +\frac{n_i\,\pii(a|s)}{n\,\piavg(a|s)}
    \rbra{
        r-\frac{\sum_{j=1}^K\!\alp^\star_j\,\pij(a|s)}{\pie(a|s)}
    }
\end{align*}
where $\balp^\star\coloneq(\alp^\star_i)_{i\in[K]}$ is a solution of $\bT\balp=\bc$. Here, $\bT\in\rr^{K\times K}$ and $\bc\in\rr^K$ have components
\begin{align*}
    T_{ij}\coloneq \expec_{\mupii}\!\sbra{\frac{\pij(A\mid S)}{\piavg(A\mid S)}},\quad
    c_i\coloneq \expec_{\mupii}\!\sbra{R\,\frac{\pie(A\mid S)}{\piavg(A\mid S)}}.
\end{align*}
The theorem states that $\bw^\star$ is an optimal solution to the variance minimization problem~\eqref{eq:var-min-problem}:
\begin{align*}
    & \underset{\bw\in(L^2)^K}{\text{minimize}}
    & &\hspace{-0.5in}\var\big[\hatJ(\pie,\bw;\D)\big]\\
    & \text{subject to}
    & &\hspace{-0.5in}\sum^K_{i=1} w_i(\omega)=r, \quad\forall \omega\in\supp\,(\mupie) 
\end{align*}
\paragraph{Variance Characterization.} We begin with a variance characterization of the generalized weighted-IPS estimator $\hatJ(\pie,\bw;\D)$ in~\eqref{eq:general-weighted-ips}. This characterization enables functional differentiation with respect to $\bw$.
\begin{lemma}
    For any weights $\bw\in(L^2)^K$ whose weighted scores $w_i(S,A,R)\,\pie(A\mid S)/\pii(A\mid S)$ have finite second moments under $\mupii$,
    \begin{align*}
        \var\big[\hatJ(\pie,\bw;\D)\big]=\sum^K_{i=1}\frac{1}{n_i}\rbra{\innp{w_i^2,\,\frac{\pie}{\pii}}_{\mupie}-\innp{w_i,\,1}^2_{\mupie}}.
    \end{align*}
    \label{lem:estimator-variance}
\end{lemma}

\begin{proof}[Proof of Lemma \ref{lem:estimator-variance}.]\label{proof:lem:estimator-variance}
As $\hatJ_{ij}(w_i;\D_i)$ are independent to each other,
\begin{align*}
    \var[\hatJ(\pie,\bw;\D)]&=\var\sbra{\sum^K_{i=1}\frac{1}{n_i}\sum^{n_i}_{j=1} \hatJ_{ij}(w_i;\D_i) }=\sum^K_{i=1} \frac{1}{n_i^2}\sum_{j=1}^{n_i}\var\sbra{\hatJ_{ij}(w_i;\D_i)}\\
    &=\sum^K_{i=1}\frac{1}{n^2_i}\sum^{n_i}_{j=1}\expec\sbra{\hatJ^{\,2}_{ij}(w_i;\D_i)}-\sum^K_{i=1}\frac{1}{n^2_i}\sum^{n_i}_{j=1}\expec\sbra{\hatJ_{ij}(w_i;\D_i)}^2.
\end{align*}

The first term is
\begin{align*}
    \sum^K_{i=1}\frac{1}{n^2_i}\sum^{n_i}_{j=1}\expec\sbra{\hatJ^{\,2}_{ij}(w_i;\D_i)}
    &=\sum^K_{i=1}\frac{1}{n_i^2}\sum_{j=1}^{n_i}\int_\Omega \frac{w_i^2(\omega)\pie^2(a|s)}{\pi_i^2(a|s)}\, p_r(r|s,a)\,\pi_i(a|s)\, p_s(s)\,d\mu(\omega)\\
    &=\sum^K_{i=1}\frac{1}{n_i}\int_\Omega \frac{w_i^2(\omega)\pie(a|s)}{\pi_i(a|s)}\, p_r(r|s,a)\,\pie(a|s)\, p_s(s)\,d\mu(\omega)\\
    &=\sum^K_{i=1}\frac{1}{n_i}\int_\Omega w_i^2(\omega)\, \frac{\pie(a|s)}{\pi_i(a|s)} \,d\mupie(\omega)\\
    &=\sum^K_{i=1}\frac{1}{n_i}\innp{w_i^2,\,\frac{\pie}{\pi_i}}_{\mupie}.
\end{align*}

The second term is
\begin{align*}
    \sum^K_{i=1}\frac{1}{n^2_i}\sum^{n_i}_{j=1}\expec\big[\hatJ_{ij}(w_i;\D_i)\big]^2
   &=\sum^K_{i=1}\frac{1}{n_i^2}\sum^{n_i}_{j=1}\rbra{\int_\Omega \frac{w_i(\omega)\pie(a|s)}{\pi_i(a|s)}\, p_r(r|s,a)\,\pi_i(a|s)\, p_s(s)\,d\mu(\omega)  }^2\\
   &=\sum^K_{i=1}\frac{1}{n_i}\rbra{\int_\Omega w_i(\omega)\, p_r(r|s,a)\,\pie(a|s)\, p_s(s)\,d\mu(\omega)}^2\\
   &=\sum^K_{i=1}\frac{1}{n_i}\rbra{\int_\Omega w_i(\omega) \,d\mupie(\omega)}^2\\
   &=\sum^K_{i=1}\frac{1}{n_i}\innp{w_i,1}^2_{\mupie}.
\end{align*}
\end{proof}
\paragraph{Proof of Optimality of $\bw^\star$.} We now prove Theorem~\ref{thm:optimalIPS}, which is restated below.
\optimalIPS*
\begin{proof}[Proof of Theorem~\ref{thm:optimalIPS}.]\label{proof:thm:optimalIPS}

Since the objective in~\eqref{eq:var-min-problem} is the variance of a linear functional of $\bw$, the variance minimization problem subject to the normalization constraint is convex. We therefore derive the first-order stationarity condition for the Lagrange-augmented functional $I[\bw,\lam]$ defined below.

Recall $\rho_r(\omega)=r$ for $\omega=(s,a,r)$. Define the functional $I[\bw, \lam]$ as
\begin{align*}
    I[\bw, \lam] &\coloneq \var\big[\hatJ(\pie, \bw; \D)\big] 
    - \innp{\lam, \sum_{i=1}^K w_i - \rho_r}_{\mupie} \\
    &= \var\big[\hatJ(\pie, \bw; \D)\big]
    - \sum_{i=1}^K \innp{\lam, w_i}_{\mupie} 
    + \innp{\lam, \rho_r}_{\mupie},
\end{align*}
where $\lam : \Omega \to \rr$ is the Lagrange multiplier enforcing the normalization constraint.

Next, for a perturbation direction $\bdel(\omega) \coloneq (\del_i(\omega))_{i \in [K]}$ and $\beps\in\rr^K$, define $I(\beps; \bdel,\bw, \lam)$ as
\begin{align*}
    I(\beps; \bdel, \bw, \lam) \coloneq I[w_1 + \eps_1 \del_1, \dots, w_K + \eps_K \del_K, \lam],
\end{align*}
which can be interpreted as the variation of $I[\bw, \lam]$ under perturbations along the direction $\bdel$.

To verify optimality, it suffices to exhibit feasible $\bw^\star$ and $\lam^\star$ such that, for every perturbation $\bdel$ with finite weighted second moments and each $i\in[K]$,
\begin{align*}
    \frac{\partial I(\beps;\bdel,\bw^\star,\lam^\star)}{\partial \eps_i}\bigg|_{\beps=0}=0.
\end{align*}
By Lemma~\ref{lem:estimator-variance}, this is equivalent to
\begin{align*}
    \innp{\del_i,\frac{2w^\star_i\pie }{n_i\pii}-\frac{2\innp{w^\star_i,1}_{\mupie}}{n_i}-\lam^\star}_{\mupie}=0.
    \numberthis\label{eq:first-order-cond}
\end{align*}
Next, we deduce $\bw^\star$ and $\lam^\star$ from~\eqref{eq:first-order-cond} and verify that the deduced $\bw^\star$ is indeed a minimizer. 

To proceed, let us define $\alp_i\coloneq\innp{w^\star_i,1}_{\mupie}$. Then,~\eqref{eq:first-order-cond} implies that, for almost every $\omega\coloneq(s,a,r)\in\supp(\mupie)$,
\begin{align*}
    \frac{2w^\star_i(\omega)\pie(a|s)}{n_i\pii(a|s)}-\frac{2\alp_i}{n_i} - \lam^\star(\omega)=0.
\end{align*}
Rearranging the above and summing over $i\in [K]$, we obtain
\begin{align}
    w^\star_i(\omega)&=\frac{\alp_i \pii(a|s)}{\pie(a|s)}+\frac{n_i\pii(a|s)}{2\pie(a|s)}\, \lam^\star(\omega) \label{eq:w-star-intermediate}\\
    &\hspace{0.5in}\implies r=\sum_{i=1}^Kw^\star_i(\omega)=\frac{\sum^K_{i=1}\alp_i \pii(a|s)}{\pie(a|s)}+\frac{\sum^K_{i=1} n_i\pii(a|s)}{2\pie(a|s)}\, \lam^\star(\omega), \nonumber
\end{align}
where we used the fact that $\sum^K_{i=1}w^\star_i(\omega)=r$ for all $\omega\in\supp\,(\mupie)$. 

Solving the above for $\lam^\star(\omega)$, 
\begin{align*}
    \lam^\star(\omega)=\frac{2r\,\pie(a|s)-2\sum^K_{i=1}\alp_i \pii(a|s)}{\sum^K_{i=1} n_i \pii(a|s)}.
\end{align*}
Plugging $\lam^\star(\omega)$ back into \eqref{eq:w-star-intermediate} results in
\begin{align*}
    w^\star_i(\omega)
    &=\frac{\alp_i\pii(a|s)}{\pie(a|s)}
    +
    \frac{n_i\,\pii(a|s)}{\sum^K_{j=1}\!n_j\pij(a|s)}
    \rbra{
        r-\frac{\sum^K_{j=1}\alp_j\pij(a|s)}{\pie(a|s)}
    },\nonumber\\
    &=
    \frac{\alp_i\pii(a|s)}{\pie(a|s)}
    +
    \frac{n_i}{n}\,\frac{\pii(a|s)}{\piavg(a|s)}
    \rbra{
        r-\frac{\sum^K_{j=1}\alp_j\pij(a|s)}{\pie(a|s)}
    }
    \numberthis\label{eq:w-star-in-proof}
\end{align*}
where one can easily verify that the omitted constraint ``$\pi_i(a|s)=0\implies w_i(\omega)=0$ for all $i\in [K]$'' is satisfied for any $\balp$.

Next, we derive a self-consistency equation for $\balp$ by plugging the above expression for $w^\star_i(\omega)$ into $\alp_i\coloneq\innp{w^\star_i,1}_{\mupie}$,
\begin{align*}
    \alp_i
    &=
    \alp_i\,\expec_{\mupie}\!\sbra{\frac{\pii(A\mid S)}{\pie(A\mid S)}}
    +\frac{n_i}{n}\,\expec_{\mupie}\!\sbra{\frac{R\,\pii(A\mid S)}{\piavg(A\mid S)}}\\
    &\qquad\qquad
    -\frac{n_i}{n}\,\sum^K_{j=1}\alp_j\,
    \expec_{\mupie}\!\sbra{\frac{\pii(A\mid S)}{\piavg(A\mid S)}\,\frac{\pij(A\mid S)}{\pie(A\mid S)}}.
    \numberthis\label{eq:selfcon}
\end{align*}
Since
\begin{align*}
    \expec_{\mupie}\!\sbra{\frac{\pii(A\mid S)}{\pie(A\mid S)}} &= \int_{\Omega} \frac{\pii(a|s)}{\pie(a|s)}\, p_r(r|s,a)\,\pie(a|s)\, p_s(s)\,d\mu(\omega) \\
    &=\int_{\Omega} p_r(r|s,a)\,\pii(a|s)\, p_s(s)\,d\mu(\omega)  \\
    &=\int_\Omega \mupii(\omega) = 1,
\end{align*}
and 
\begin{align*}
    T_{ij}&\coloneq\expec_{\mupii}\!\sbra{\frac{\pij(A\mid S)}{\piavg(A\mid S)}}=\expec_{\mupie}\!\sbra{\frac{\pii(A\mid S)}{\piavg(A\mid S)}\,\frac{\pij(A\mid S)}{\pie(A\mid S)}}\\
    c_i&\coloneq\expec_{\mupii}\!\sbra{\frac{R\,\pie(A\mid S)}{\piavg(A\mid S)}}=\expec_{\mupie}\!\sbra{\frac{R\,\pii(A\mid S)}{\piavg(A\mid S)}}
\end{align*}
The self-consistency equation~\eqref{eq:selfcon} can be written as $\bT\balp=\bc$ with $\bT$ and $\bc$ given in~\eqref{eq:Tc}.

Finally, by Lemma~\ref{lem:self-consistency-as-projection}, $\bT\balp=\bc$ admits a solution. Hence, the corresponding weights $\bw$ satisfy the first-order optimality condition~\eqref{eq:first-order-cond} and therefore solve the variance minimization convex program~\eqref{eq:var-min-problem}.
\end{proof}

\subsection{Proof of Unbiasedness and Asymptotic Variance Optimality of $\hatJ_\oips$}
In this section, we prove that $\hatJ_\oips$ is unbiased and asymptotically normal, with asymptotic variance $n\var\big[\hatJ(\pie,\bw^\star;\D)\big]$, which is optimal within $\wipsclass$.

\optimalIPSasymptotics*

\begin{proof}[Proof of Theorem~\ref{thm:optimalIPSasymptotics}]\label{proof:thm:optimalIPSasymptotics}
    We first prove the unbiasedness of $\hatJ_\oips(\pie;\D)$ when the logging policies are known.

    \paragraph{Unbiasedness.} We begin by rewriting the expectation of $\hatJ_\oips(\pie;\D)$ as an iterated expectation, conditioning on the data $\D^{(-z)}$ used to estimate the nuisance parameter $\balp$:
    \begin{align*}
        \expec\sbra{\hatJ_\oips(\pie;\D)} &= \frac{1}{n}\sum_{z=1}^Z n^{(z)} \,\expec\sbra{\hatJ(\pie,\hatbalp^{(z)};\D^{(z)})}\\
        &=\frac{1}{n}\sum_{z=1}^Z n^{(z)}\,\expec\sbra{\expec\big[\hatJ(\pie,\hatbalp^{(z)};\D^{(z)})\,\big|\,\D^{(-z)} \big]}
    \end{align*}

    It therefore suffices to show that
    \begin{align*}
        \expec\big[\hatJ(\pie,\hatbalp^{(z)};\D^{(z)})\big| \D^{(-z)} \big] = J(\pie),
    \end{align*}
    for each fold $z$. To see this, let $\hatbw^{(z)}$ be defined by
    \begin{align*}
        \hatw_i^{(z)}(s,a,r)\coloneq \frac{\hatalp_i^{(z)}\pii(a|s)}{\pie(a|s)} + \frac{n_i^{(z)}\pii(a|s) }{\sum_{j=1}^K n_j^{(z)}\pij(a|s)}\rbra{r - \frac{\sum_{k=1}^K \hatalp_k^{(z)}\pik(a|s)}{\pie(a|s)}}
    \end{align*}
    Then, conditional on $\hatbalp^{(z)}$, these weights satisfy the normalization condition $\sum_{i=1}^K \hatw_i^{(z)}(s,a,r)=r$ and the dominance condition $\pii(a|s)=0 \implies \hatw_i^{(z)}(s,a,r)=0$.

    Moreover, $\hatJ(\pie,\hatbalp^{(z)};\D^{(z)})$ in~\eqref{eq:hatJ-control-variate-form} can be rewritten in the canonical generalized weighted-IPS form~\eqref{eq:general-weighted-ips}, with $\hatbw^{(z)}$ as its weights:
    \begin{align*}
        \hatJ(\pie,\hatbalp^{(z)};\D^{(z)})=\sum_{i=1}^K \frac{1}{n_i^{(z)}} \sum_{(S,A,R)\in \D_i^{(z)}} \hatw_i^{(z)}(S,A,R)\, \frac{\pie(A\mid S)}{\pii(A\mid S)}.
    \end{align*}
    Since $\hatbw^{(z)}$ satisfies the normalization and dominance conditions conditional on $\hatbalp^{(z)}$, Theorem~\ref{thm:unbiasedness} implies that $\hatJ(\pie,\hatbalp^{(z)};\D^{(z)})$ is unbiased conditional on $\hatbalp^{(z)}$, equivalently conditional on $\D^{(-z)}$.

    \paragraph{Asymptotic Normality.} Next, we show that
    \begin{align*}
        \sqrt{n'}\rbra{\hatJ_\oips(\pie;\D) - J(\pie)} \overset{d}{\to} \N\rbra{0, n\,\var[\hatJ(\pie,\vw^\star;\D)]}.
    \end{align*}
    We work with known logging policies under Assumption~\ref{as:regularity}: $K$ and $Z$ are fixed, the logger-wise folds are balanced up to rounding, the fitted scores converge in probability to the oracle score in each $L^2(\mupii)$, and all fitted and oracle scores have finite second moments, including over training randomness.

    We rewrite our infeasible optimal estimator $\hatJ(\pie,\balp^\star;\D)$ in~\eqref{eq:hatJ-control-variate-form} in the form considered by~\citet{kallus2021optimal}:
    \begin{align*}
        \hatJ(\pie,\balp^\star;\D)= \frac{1}{n}\sum_{i=1}^K \sum_{(S,A,R)\in\D_i}\sbra{\frac{\pie(A\mid S)}{\piavg(A\mid S)}\rbra{R - \frac{g_{\balp^\star}}{\pie}(S,A)} + \frac{g_{\balp^\star}}{\pie}(S,\pie)}.
    \end{align*}

    The normal equation gives the oracle summand mean $J(\pie)$ under every logger. Grouping the oracle observations into $m$ independent blocks, each containing $n_i$ observations from logger $i$, the i.i.d. central limit theorem gives the displayed limit.

    Conditional independence and the assumed $L^2$ score convergence make each fold's contribution to the centered error relative to the oracle, scaled by $\sqrt{n'}$, $o_p(1)$ by conditional Chebyshev. With known policies, the conditional bias is zero. Since $Z$ is fixed and fold weights are sample proportions, summing the contributions and applying Slutsky's theorem gives the desired asymptotic normality result. If the logging policies are instead estimated by normalized policies fitted on the training folds, with positive mixtures, the same argument applies whenever each fold mean has conditional bias $o_p(n'^{-1/2})$; this is the condition in Remark~\ref{rem:estimated}, and it is not implied by $\sqrt{n}$-consistency of the estimated propensities. The same argument applies to DR scores under the corresponding moment, score-convergence, and bias conditions.
\end{proof}

\subsection{Additional Proofs}
\begin{lemma}[Normal-Equation Interpretation of $\bT\balp=\bc$]
    For $\bT$ and $\bc$ defined in~\eqref{eq:Tc}, their entries can be written as
    \begin{align*}
        T_{ij}=\big\langle\pii,\pij\big\rangle_{\mulog},\quad c_i=\big\langle q\,\pie,\pii \big\rangle_{\mulog},
    \end{align*}
    where $\mulog$ is given by 
    \begin{align*}
        \frac{d\mulog}{d\mu_{s,a}}(s,a)\coloneq\frac{p_s(s)}{\piavg(s,a)},
    \end{align*}
    with $\mu_{s,a}$ denoting the base measure over $\calS\times\A$. 

    Consequently, $\bT\balp=\bc$ is the normal equation for the orthogonal projection of $q\,\pie$ onto $\V\coloneq\mathrm{span}\big(\{\pii\}_{i\in[K]}\big)$. In particular, $\bT\balp=\bc$ admits a solution, and any solution $\balp^\star$ defines the orthogonal projection of $q\,\pie$ onto $\V$ through $\sum_{i\in [K]}\alp_i^\star\pii$.
    \label{lem:self-consistency-as-projection}
\end{lemma}

\begin{proof}[Proof of Lemma~\ref{lem:self-consistency-as-projection}]
\label{proof:thm:self-consistency-as-projection}
    Recall the definitions of $\bT$ and $\bc$ in~\eqref{eq:Tc}:
    \begin{align*}
        T_{ij}\coloneq \expec_{\mupii}\!\sbra{\frac{\pij(A\mid S)}{\piavg(A\mid S)}},\ \  
        c_i\coloneq \expec_{\mupii}\!\sbra{R\,\frac{\pie(A\mid S)}{\piavg(A\mid S)}}.
    \end{align*}

    The first identity follows by direct computation:
    \begin{align*}
        T_{ij}
        &= \expec_{\mupii}\!\sbra{\frac{\pij(A\mid S)}{\piavg(A\mid S)}}\\
        &= \int_\Omega \frac{\pii(a|s)\pij(a|s)}{\piavg(a|s)}\,p_r(r|s,a)\,p_s(s)\,d\mu(\omega)\\
        &=\int_{\calS\times\A}\pii(a|s)\,\pij(a|s)\,\frac{p_s(s)}{\piavg(a|s)}\,d\mu_{s,a}(s,a)\\
        &=\big\langle\pii,\pij\big\rangle_{\mulog}.
    \end{align*}
    Similarly, for $c_i$, we have
    \begin{align*}
        c_i
        &= \expec_{\mupii}\!\sbra{R\,\frac{\pie(A\mid S)}{\piavg(A\mid S)}}\\
        &= \expec_{\mupii}\!\sbra{\expec_{\mupii}\!\sbra{R\,\frac{\pie(A\mid S)}{\piavg(A\mid S)}\,\middle|\, S, A
        }}\\
        &=\expec_{\mupii}\!\sbra{\expec_{\mupii}\!\sbra{R\,\middle|\, S, A
        }\,\frac{\pie(A\mid S)}{\piavg(A\mid S)}}\\
        &=\expec_{\mupii}\!\sbra{q(S,A)\,\frac{\pie(A\mid S)}{\piavg(A\mid S)}} \\
        &=\int_\Omega q(s,a)\, \frac{\pie(a|s)\,\pii(a|s)}{\piavg(a|s)}\, p_r(r|s,a)\,p_s(s)\,d\mu(\omega)\\
        &=\int_{\calS\times\A} q(s,a)\,\pie(a|s)\,\pii(a|s)\frac{p_s(s)}{\piavg(a|s)}\,d\mu_{s,a}(s,a)\\
        &=\big\langle q\,\pie,\pii \big\rangle_{\mulog}.
    \end{align*}

    It remains to show that $\bT\balp=\bc$ is the normal equation for the orthogonal projection. Let
    \begin{align*}
        \V\coloneq \mathrm{span}\big(\{\pii\}_{i\in[K]}\big)\subset L^2(\mulog).
    \end{align*}
    Since $\V$ is finite-dimensional, it is closed. Hence, the orthogonal projection of $q\,\pie$ onto $\V$ exists. Write this projection as
    \begin{align*}
        g_{\balp^\star}\coloneq \sum_{k\in[K]}\alp_k^\star\pik.
    \end{align*}
    By the characterization of orthogonal projections, $g_{\balp^\star}$ satisfies
    \begin{align*}
        \big\langle q\,\pie-g_{\balp^\star},\pii\big\rangle_{\mulog}=0,
        \quad \forall i\in[K].
    \end{align*}
    Expanding this condition gives
    \begin{align*}
        \big\langle q\,\pie,\pii\big\rangle_{\mulog}
        =
        \sum_{k\in[K]}\alp_k^\star\big\langle \pik,\pii\big\rangle_{\mulog},
        \quad \forall i\in[K].
    \end{align*}
    By the identities proved above, this system is exactly $\bT\balp^\star=\bc$. Therefore, $\bT\balp=\bc$ admits a solution, and any solution $\balp^\star$ defines the orthogonal projection of $q\,\pie$ onto $\V$ through $g_{\balp^\star}=\sum_{i\in[K]}\alp_i^\star\pii$.
\end{proof}

\LemVarianceGap

\begin{proof}[Proof of Lemma~\ref{lem:variance-gap}]
\label{proof:lem:variance-gap}

Let $Q_f=\pie(R-f)/\piavg+f(S,\pie)$ and $V(f)=\var_{\mupiavg}[Q_f]$. Independence across samples and the variance decomposition over loggers give
\begin{align*}
    \var[\hatJ_\Gamma(\pie,f;\D)]
    =\frac{1}{n^2}\sum_i n_i\var_{\mupii}[Q_f]
    =\frac{1}{n}\rbra{V(f)-\sum_i\frac{n_i}{n}m_i(f)^2}.
\end{align*}
Here $m_i(f)=\expec_{\mupii}[Q_f]-J(\pie)$, which agrees with the definition in the lemma, the second equality uses $\expec_{\mupiavg}[Q_f]=J(\pie)$, and $m_i(q)=0$.
For the remainder of the proof, all variances, covariances, and expectations are taken with respect to $\mupiavg$, so we omit the corresponding subscript.

Using $\var[X]-\var[Y]=\var[X-Y]+2\cov[X-Y,Y]$, we obtain
\begin{align*}
    &\frac{V(f)-V(q)}{n}\\
    &=\frac{1}{n}\var\!\sbra{\frac{\pie(A\mid S)}{\piavg(A\mid S)}\big(q(S,A)-f(S,A)\big)-\big(q(S,\pie)-f(S,\pie)\big)}\\
    &\quad+\frac{2}{n}\cov\!\left[\begin{aligned}
      &\frac{\pie(A\mid S)}{\piavg(A\mid S)}\big(q(S,A)-f(S,A)\big)-\big(q(S,\pie)-f(S,\pie)\big),\\
      &\frac{\pie(A\mid S)}{\piavg(A\mid S)}\big(R-q(S,A)\big)+q(S,\pie)
    \end{aligned}\right].
\end{align*}
We now show that the covariance term is zero. It decomposes into
\begin{align*}
    &\cov\sbra{
        \frac{\pie(A\mid S)}{\piavg(A\mid S)}\big(q(S,A)-f(S,A)\big)-\big(q(S,\pie)-f(S,\pie)\big),
        \frac{\pie(A\mid S)}{\piavg(A\mid S)}\big(R-q(S,A)\big)}\\
    &\hspace{1in}+
    \cov\sbra{
        \frac{\pie(A\mid S)}{\piavg(A\mid S)}\big(q(S,A)-f(S,A)\big)-\big(q(S,\pie)-f(S,\pie)\big),
        q(S,\pie)}.
\end{align*}

The first covariance vanishes because the first argument is measurable with respect to $(S,A)$, while
\begin{align*}
    \expec\!\sbra{R-q(S,A)\mid S,A}=0.
\end{align*}
Thus, by iterated expectation, its covariance with the centered residual term is zero.

For the second covariance, note that $q(S,\pie)$ depends only on $S$. Moreover, conditional on $S$,
\begin{align*}
    \expec\!\sbra{
    \frac{\pie(A\mid S)}{\piavg(A\mid S)}\big(q(S,A)-f(S,A)\big)
    \,\middle|\, S}
    =
    q(S,\pie)-f(S,\pie).
\end{align*}
Hence,
\begin{align*}
    \expec\!\sbra{
    \frac{\pie(A\mid S)}{\piavg(A\mid S)}\big(q(S,A)-f(S,A)\big)-\big(q(S,\pie)-f(S,\pie)\big)
    \,\middle|\, S}
    =0.
\end{align*}
Since the second argument $q(S,\pie)$ is a function only of $S$, the second covariance also vanishes. Therefore, the covariance term is zero, and the difference of mixture variances reduces to
\begin{align*}
    &\frac{V(f)-V(q)}{n}\\
    &=\frac{1}{n}\var\!\sbra{\frac{\pie(A\mid S)}{\piavg(A\mid S)}\big(q(S,A)-f(S,A)\big)-\big(q(S,\pie)-f(S,\pie)\big)}\geq 0.
\end{align*}

We proceed further to put the above expression into the claimed form. Define
\begin{align*}
    X\coloneq \frac{\pie(A\mid S)}{\piavg(A\mid S)}\big(q(S,A)-f(S,A)\big),
    \qquad
    Y\coloneq q(S,\pie)-f(S,\pie).
\end{align*}
Under $\mupiavg$, the conditional distribution of $A$ given $S$ is $\piavg(\cdot\mid S)$. Hence,
\begin{align*}
    \expec\sbra{X\mid S}
    &= \int_{\A}\frac{\pie(a\mid S)}{\piavg(a\mid S)}\big(q(S,a)-f(S,a)\big)\piavg(a\mid S)\,d\mu_a(a)\\
    &= \int_{\A}\pie(a\mid S)\big(q(S,a)-f(S,a)\big)\,d\mu_a(a)\\
    &= q(S,\pie)-f(S,\pie)
    =Y.
\end{align*}
Therefore, the last display can be written as
\begin{align*}
    \frac{V(f)-V(q)}{n}=
    \frac{1}{n}\var\!\sbra{X-\expec[X\mid S]}.
\end{align*}
By the law of total variance,
\begin{align*}
    \var\sbra{X}=
    \expec\sbra{\var\sbra{X\mid S}}+\var\sbra{\expec\sbra{X\mid S}}.
\end{align*}
Since $\expec[X\mid S]=Y$, this gives
\begin{align*}
    \var\sbra{X-\expec[X\mid S]}=
    \var\sbra{X}-\var\sbra{Y}\geq 0.
\end{align*}
Substituting the definitions of $X$ and $Y$, and noting that the marginal distribution of $S$ under $\mupiavg$ is $p_s$, we obtain
\begin{align*}
    &\frac{V(f)-V(q)}{n}\\
    &=\frac{1}{n}\rbra{\var\sbra{\frac{\pie(A\mid S)}{\piavg(A\mid S)}\big(q(S,A)-f(S,A)\big)}-\var_{p_s}\big[q(S,\pie)-f(S,\pie)\big]}\geq 0.
\end{align*}
Subtracting $n^{-1}\sum_i(n_i/n)m_i(f)^2$ proves the claimed identity. Nonnegativity follows since the result equals $n^{-1}\sum_i(n_i/n)\var_{\mupii}[X-Y]$. Here we use $\sum_i(n_i/n)\cov_{\mupii}[X-Y,Q_q]=\cov_{\mupiavg}[X-Y,Q_q]=0$, where the first equality holds because $\expec_{\mupii}[Q_q]=J(\pie)$ for every $i$.
\end{proof}

\ThmVarianceGap*
\begin{proof}[Proof of Theorem~\ref{thm:variance-gap}]
\label{proof:thm:variance-gap}
    First, rewrite $\hatJ(\pie,\balp^\star;\D)$ in~\eqref{eq:hatJ-control-variate-form} as
    \begin{align*}
        \hatJ(\pie,\balp^\star;\D)= \frac{1}{n}\sum_{i=1}^K \sum_{(S,A,R)\in\D_i}\sbra{\frac{\pie(A\mid S)}{\piavg(A\mid S)}\rbra{R - f_{\balp^\star}(S,A)} + f_{\balp^\star}(S,\pie)},
    \end{align*}
    where $f_{\balp^\star}\coloneq g_{\balp^\star}/\pie = \sum_{i=1}^K \alp^\star_i \pii/\pie$. Hence, $\hatJ(\pie,\balp^\star;\D)=\hatJ_\Gamma(\pie,f_{\balp^\star};\D)$.

    The normal equation gives $\expec_{\mupii}[\pie(q-f_{\balp^\star})/\piavg]=0$ for every $i$. Also, weighting the $i$-th normal equation by $n_i/n$ and using $\sum_i\frac{n_i}{n}T_{ij}=1$ and $\sum_i\frac{n_i}{n}c_i=J(\pie)$ gives $\sum_j\alp^\star_j=J(\pie)$, so $m_i(f_{\balp^\star})=0$. Applying Lemma~\ref{lem:variance-gap} therefore gives
    \begin{align*}
        \var\big[\hatJ(\pie,\balp^\star;\D) \big] - \var\big[ \hatJ^\star_\Gamma(\pie;\D)\big]&\\
        &\hspace{-1.7in}=\frac{1}{n}\rbra{\var_{\mupiavg}\!\sbra{\frac{\pie(A\mid S)}{\piavg(A\mid S)}\big(q(S,A)-f_{\balp^\star}(S,A)\big)}-\var_{p_s}\big[q(S,\pie)-f_{\balp^\star}(S,\pie)\big]}\\
        &\hspace{-1.7in}=\frac{1}{n}\rbra{\var_{\mupiavg}\!\sbra{\frac{\pie(A\mid S)}{\piavg(A\mid S)}\big(q(S,A)-f_{\balp^\star}(S,A)\big)}-\var_{p_s}\big[q(S,\pie)\big]}.
    \end{align*}
    The last equality follows because
    \begin{align*}
        f_{\balp^\star}(S,\pie)
        =\int_\A \sum_{i=1}^K\alp_i^\star \pii(a|s)\,d\mu_a(a)
        = \sum_{i=1}^K\alp_i^\star,
    \end{align*}
    which is deterministic and therefore does not contribute to the variance.

    It remains to rewrite the first variance term in terms of the norm $\|\cdot\|_{\mulog}$. First, observe that the term inside $\var_{\mupiavg}$ has mean zero:
    \begin{align*}
        \expec_{\mupiavg}\sbra{\frac{\pie(A\mid S)}{\piavg(A\mid S)}\big(q(S,A)-f_{\balp^\star}(S,A)\big)}&=\expec_{\mupie}\big[q(S,A)-f_{\balp^\star}(S,A)\big] \\
        &=\expec_{\mupie}\big[q(S,A)\big] - \expec_{\mupie}\!\sbra{\sum_{i=1}^K \alp^\star_i \frac{\pii(A|S)}{\pie(A|S)}}\\
        &=J(\pie) - \sum_{i=1}^K\alp^\star_i=0.
    \end{align*}
    Here, the last equality follows from the definition $\alp^\star_i=\innp{w_i,1}_{\mupie}$ in the proof of Theorem~\ref{thm:optimalIPS} and the normalization condition $\sum_{i\in[K]} w_i(s,a,r)=r$, which imply
    \begin{align*}
        \sum_{i\in[K]}\alp^\star_i
        = \Big\langle\sum_{i\in [K]} w_i,1\Big\rangle_{\mupie}
        = \expec_{\mupie}\sbra{R}
        = J(\pie).
    \end{align*}

    Therefore,
    \begin{align*}
        \var_{\mupiavg}\!\sbra{\frac{\pie(A\mid S)}{\piavg(A\mid S)}\big(q(S,A)-f_{\balp^\star}(S,A)\big)}
        &= \expec_{\mupiavg}\!\sbra{\frac{\pie^2(A\mid S)}{\piavg^2(A\mid S)}\big(q(S,A)-f_{\balp^\star}(S,A)\big)^2}\\
        &=\expec_{\mupiavg}\!\sbra{\frac{\big(q(S,A)\pie(A\mid S)-g_{\balp^\star}(S,A)\big)^2}{\piavg^2(A\mid S)}}\\
        &=\big\|q\,\pie - g_{\balp^\star}\big\|^2_{\mulog}\\
        &=\inf_{\balp\in\rr^K}\big\|q\,\pie - g_{\balp}\big\|^2_{\mulog},
    \end{align*}
    where the last equality follows from the fact that $g_{\balp^\star}$ is the orthogonal projection of $q\,\pie$ with respect to the norm $\|\cdot\|_{\mulog}$. This proves the theorem.
\end{proof}

\section{Computational Complexity and Scalability}
\label{app:complexity}
In this section, we analyze the computational complexity of $\hatJ_{\oips}$, especially when the number of loggers is large ($K\approx100$). Although restricted-optimal coefficients require solving a linear system with technique matrix $\bT \in \rr^{K \times K}$, we show both theoretically and empirically that the added cost is modest for practical $K$. In this appendix, $N$ denotes the total sample size, written $n$ in the main text.

\subsection{Theoretical Complexity Analysis}

Computing $\hatJ_{\oips}$ within cross-fitting has two main steps: (i) estimating the technique matrix $\bT$ and contribution vector $\bc$, and (ii) solving for $\balp = \bT^{-1}\bc$ (or its pseudoinverse analogue).

\begin{enumerate}[leftmargin=*,label=(\roman*)]
    \item \textbf{Matrix construction:} Constructing $\bT$ aggregates propensity terms over $N$ samples and $K$ loggers, yielding $O(NK)$ complexity. This matches the dominant complexity class of balanced IPS ($\hatJ_{\bips}$), which also computes aggregate propensity statistics (e.g., $\sum_k n_k \pi_k$) per sample.
    \item \textbf{Linear solve / pseudoinversion:} Solving $\bT\balp = \bc$ has worst-case complexity $O(K^3)$ with standard dense linear algebra (e.g., LU or SVD).
\end{enumerate}

The total time is $O(Z(NK+K^3))$, which is $O(NK+K^3)$ for fixed $Z$. Accumulating the per-fold moments in one pass uses $O(ZK^2)$ additional memory, while storing all cross-logger propensities uses $O(NK)$. These counts exclude fitting nuisance models such as propensity models. We compute $\balp$ with the SVD-based Moore--Penrose pseudoinverse, which also handles a rank-deficient $\bT$; this occurs in the runtime study below, where all loggers are mixtures of the deterministic and uniform policies.

\subsection{Empirical Runtime Analysis}

To validate practicality, we measured runtime on the \textit{Letter} dataset ($N = 20{,}000$, $26$ actions) with $K=10$ and $K=100$ loggers. Experiments ran on an Apple MacBook Air M1 (base model, 8GB RAM). Logging policies were generated by mixing the deterministic policy with a uniform policy. We compared Naive IPS, Balanced IPS, Weighted IPS~\citep{agarwal2017effective}, and our Optimal IPS. Methods requiring sample-dependent weights (Weighted IPS and Optimal IPS) used $Z=5$ cross-fitting folds.

Table~\ref{tab:runtime_results} reports mean inference time per trial (milliseconds) over 200 independent runs. Each $\pm$ value is the half-width of a $95\%$ confidence interval based on the normal approximation.

\begin{table}[h]
\centering
\caption{Comparison of Relative RMSE and wall-clock Time (in milliseconds) on the Letter dataset with $N=20{,}000$. We compare settings with $K=10$ and $K=100$ logging policies. Our method achieves the lowest error while maintaining a runtime comparable to the existing Weighted IPS baseline.}
\label{tab:runtime_results}
\vspace{0.2cm}
\resizebox{0.8\textwidth}{!}{%
\begin{tabular}{l|cc|cc}
\toprule
\multirow{2}{*}{\textbf{Estimator}} & \multicolumn{2}{c|}{\textbf{$K=10$ Loggers}} & \multicolumn{2}{c}{\textbf{$K=100$ Loggers}} \\
 & \textbf{Rel-RMSE} & \textbf{Time (ms)} & \textbf{Rel-RMSE} & \textbf{Time (ms)} \\
\midrule
Naive IPS ($\hatJ_{\ips}$) & 0.0113 & $0.16 \pm 0.02$ & 0.0104 & $0.18 \pm 0.02$ \\
Balanced IPS ($\hatJ_{\bips}$) & 0.0133 & $0.13 \pm 0.02$ & 0.0087 & $0.51 \pm 0.01$ \\
Weighted IPS ($\hatJ_{\wips}$) & 0.0122 & $2.16 \pm 0.03$ & 0.0075 & $15.49 \pm 0.43$ \\
\textbf{Optimal IPS ($\hatJ_{\oips}$)} & \textbf{0.0045} & $2.52 \pm 0.05$ & \textbf{0.0045} & $20.06 \pm 0.13$ \\
\bottomrule
\end{tabular}%
}
\end{table}

\textbf{Observations.}
\begin{itemize}[leftmargin=*]
    \item \textbf{Scalability:} Even at $K=100$, Optimal IPS takes about 20ms per trial. This is slower than heuristic balanced IPS but still inexpensive for offline evaluation.
    
    \item \textbf{Comparison with Weighted IPS:} Runtime is comparable to Weighted IPS. This is expected because both methods use cross-fitting and dataset passes to estimate nuisance quantities (variance terms for Weighted IPS, technique matrix for Optimal IPS). The linear solve adds limited overhead relative to those passes.
    
    \item \textbf{Efficiency--Compute Trade-off:} For a small runtime increase (milliseconds), Optimal IPS reduces Relative-RMSE by about $40\%$ versus Weighted IPS and about $48\%$ versus Balanced IPS in the $K=100$ setting.
\end{itemize}

\paragraph{Practical Takeaway.}
These numbers should be interpreted in the context of the full OPE pipeline. In our runs, end-to-end processing, including data loading, preprocessing, and fitting logger models, dominates total cost. Against that baseline, the additional time for $\hatJ_{\oips}$ remains a moderate fraction of wall-clock time rather than a bottleneck, while providing substantial accuracy gains.

\subsection{Large-$N$ Synthetic Experiment}\label{app:large-n}

We use the synthetic data generator of the Open Bandit Pipeline (OBP, version 0.5.7)~\citep{saito2021open} with $10$ actions, five-dimensional Gaussian contexts, and binary rewards from a logistic model with fixed coefficients. The evaluation policy chooses the action with the highest conditional mean reward. Each of the five known logging policies mixes the evaluation policy with the uniform policy, with uniform-exploration probabilities $0.05$, $0.25$, $0.50$, $0.75$, and $0.95$, and all strata have the same size. For each $N\in\{10^4,10^5,10^6\}$, we draw contexts, actions, and rewards with $20$ independent random seeds. The ground-truth value is the evaluation policy's conditional mean reward averaged over the sampled contexts. Weighted IPS and $\hatJ_\oips$ use five-fold cross-fitting. Confidence intervals are computed as in \S\ref{sec:experiment}, with the $20$ seeds as the resampled units.

\begin{table}[ht]
\centering
\caption{Relative-RMSE ($\times10^{-3}$) in the large-$N$ synthetic experiment. Parentheses show $95\%$ bootstrap CIs.}
\label{tab:large-n-error}
\small
\setlength{\tabcolsep}{5pt}
\begin{tabular}{lccc}
\toprule
Estimator & $N=10^4$ & $N=10^5$ & $N=10^6$ \\
\midrule
IPS & \shortstack{13.0\\(9.71, 16.2)} & \shortstack{3.92\\(2.88, 4.84)} & \shortstack{1.40\\(0.941, 1.79)} \\[3pt]
bIPS & \shortstack{6.14\\(4.32, 7.94)} & \shortstack{2.15\\(1.53, 2.72)} & \shortstack{0.729\\(0.555, 0.904)} \\[3pt]
wIPS & \shortstack{3.83\\(2.77, 4.82)} & \shortstack{1.85\\(1.23, 2.43)} & \shortstack{0.544\\(0.370, 0.718)} \\[3pt]
oIPS & \shortstack{\textbf{2.70}\\(1.89, 3.43)} & \shortstack{\textbf{0.775}\\(0.570, 0.948)} & \shortstack{\textbf{0.199}\\(0.144, 0.251)} \\
\bottomrule
\end{tabular}
\end{table}

\begin{table}[ht]
\centering
\caption{Mean runtime of each estimator in milliseconds, and the largest peak resident memory over $20$ seeds. Memory is measured for a single process that generates the data and runs all four estimators.}
\label{tab:large-n-runtime}
\small
\begin{tabular}{rccccc}
\toprule
$N$ & IPS & bIPS & wIPS & oIPS & Peak memory (MiB) \\
\midrule
$10^4$ & 0.178 & 0.161 & 4.626 & 6.000 & 124.652 \\
$10^5$ & 0.994 & 1.220 & 15.372 & 25.435 & 164.668 \\
$10^6$ & 18.053 & 23.462 & 162.890 & 346.201 & 584.516 \\
\bottomrule
\end{tabular}
\end{table}

We ran these measurements on an x86-64 Linux machine with eight allocated CPU cores and $12$ GiB of RAM. Runtimes exclude data generation and the computation of propensities; peak memory includes the stored cross-logger propensities.

\section{Related Work}\label{app:related-work}
OPE evaluates counterfactual policies from logged data~\citep{precup2000eligibility,bottou2013counter}. Although classical treatments focus on a single logger, many real applications aggregate data from multiple behavior policies, such as parallel A/B systems, leading to a stratified design in which logger identity defines strata~\citep{agarwal2017effective,he2019off,kallus2021optimal,chen2020infinite,wooldridge2001asymptotic}.

\paragraph{OPE with Multiple Loggers.}
\citet{agarwal2017effective} initiated the multi-logger OPE literature with MIS-inspired IPS estimators~\citep{veach1995optimally,cornuet2012adaptive,elvira2019generalized} and showed that no single estimator uniformly dominates across instances. \citet{kallus2021optimal} formalized this setting, derived the semiparametric efficiency bound, and proposed doubly robust estimators that asymptotically attain it. Related directions include off-policy learning with multiple loggers~\citep{he2019off}, infinite-horizon analysis~\citep{chen2020infinite}, and generalization under selection bias~\citep{hatt2022generalizing}. \citet{liu2025efficient} studied multi-policy evaluation in RL via behavior-policy design.

\paragraph{Multiple Importance Sampling.}
MIS was originally developed to combine multiple proposal distributions in Monte Carlo rendering~\citep{veach1995optimally,owen2000safe,elvira2019generalized,kondapaneni2019optimal}. The balance and power heuristics~\citep{veach1995optimally} provide robust performance under nonnegative weights, and later work studied broader weighting families~\citep{elvira2019generalized}. \citet{kondapaneni2019optimal} derived variance-optimal MIS weights that may be negative, yielding further variance reduction in some regimes. Our method adapts this variance-minimization perspective to multi-logger OPE.

\paragraph{Broader Context.} Beyond the multi-logger setting, there is extensive work on variance control for IPS in standard OPE. Classical IPS traces back to Horvitz--Thompson~\citep{horvitz1952}. Subsequent work studied weight truncation~\citep{ionides2008} and doubly robust estimation in contextual bandits~\citep{dudik2014doubly}. More recent lines include self-normalization and implicit-exploration objectives~\citep{swaminathan2015sn,kuzborskij2021confident}, transformations yielding sub-Gaussian concentration~\citep{metelli2021subgaussian}, and pessimistic regularization through PAC-Bayesian analysis~\citep{aouali2024unified}. These methods are complementary to, rather than replacements for, our multi-logger focus.

\section{Broader Impacts}
\label{app:broader_impacts}
This work is methodological research on off-policy evaluation with multiple logging policies. Its main positive impact is to make retrospective policy evaluation more reliable in domains where direct online testing may be costly or risky, such as recommender systems, healthcare decision support, and advertising. The main potential negative impacts arise from downstream use: more accurate OPE could still be used to justify the deployment of policies trained on biased, privacy-sensitive, or otherwise unrepresentative logs, and an incorrectly specified evaluation pipeline could give false confidence in a harmful policy. These risks are not unique to our estimator, but they motivate using it together with domain review, data-governance checks, privacy controls, fairness and subgroup analyses, and prospective validation before deployment in high-stakes settings.

\end{document}